\documentclass[11pt]{article}

\usepackage[final]{acl}

\usepackage{times}
\usepackage{latexsym}

\usepackage[T1]{fontenc}

\usepackage[utf8]{inputenc}

\usepackage{microtype}

\usepackage{inconsolata}

\usepackage{graphicx}

\usepackage{amsmath}
\usepackage{amssymb}
\usepackage{mathtools}
\usepackage{amsthm}
\usepackage{multirow}
\usepackage{enumitem}
\usepackage{xspace}
\usepackage{natbib}
\usepackage[section]{placeins}
\usepackage[table]{xcolor}
\usepackage{tabularx}
\usepackage{makecell}
\usepackage{float}
\usepackage{tcolorbox}
\usepackage{fvextra}
\usepackage{mdframed}
\usepackage[capitalize,noabbrev]{cleveref}
\usepackage{booktabs}
\usepackage{colortbl}

\newcolumntype{M}{>{\raggedright\arraybackslash}p{2.8cm}}
\newcolumntype{Y}{>{\centering\arraybackslash}X}

\title{Automatic Configuration of LLM Post-Training Pipelines}

\author{
  Channe Chwa, Xinle Wu, Yao Lu \\
  National University of Singapore \\
  \texttt{\{channe,wuxl,luyao\}@comp.nus.edu.sg}
}

\begin{document}
\maketitle
\begin{abstract}
LLM post-training pipelines that combine supervised fine-tuning and reinforcement learning are difficult to configure under realistic compute budgets: the configuration space is high-dimensional and heterogeneous, stages are strongly coupled, and each end-to-end evaluation is expensive. We propose \emph{AutoPipe}, a budget-aware two-phase framework for configuration selection in LLM post-training. Offline, AutoPipe learns a dataset-conditioned learning-to-rank surrogate from historical runs, capturing within-dataset preferences and providing transferable guidance toward promising regions of the configuration space. Online, for a new dataset, AutoPipe uses the offline guidance to steer Bayesian optimization and models dataset-specific deviations with a Gaussian-process residual surrogate. To reduce evaluation cost, each trial is early-stopped and scored by a learned predictor that maps early training signals to a low-cost proxy for final post-training performance. Experiments on biomedical reasoning tasks show that AutoPipe achieves performance comparable to full-fidelity HPO baselines at a fraction of their online cost, while outperforming offline-only approaches and early-stopped HPO variants.
\end{abstract}

\section{Introduction}

Large language models (LLMs) have become a central paradigm in modern AI, delivering strong generalization across language understanding, generation, and reasoning tasks~\citep{lewis2020retrieval,schick2023toolformer,singhal2023large}. This progress is commonly enabled by a two-stage training recipe: \emph{pre-training} on large-scale corpora to acquire broad linguistic and world knowledge, followed by \emph{post-training} to adapt and align a pretrained base model to downstream domains, user preferences, and task-specific evaluation criteria. In many practical settings, post-training is implemented as a multi-stage pipeline that combines supervised fine-tuning (SFT), which provides a task-adapted initialization, with reinforcement learning (RL), which further optimizes behavior against downstream objectives~\citep{wang2024survey}.

While effective, post-training pipelines are difficult to configure: practitioners must jointly choose a base model together with a heterogeneous collection of stage-specific hyperparameters, and interactions across stages can make the resulting performance landscape highly nontrivial. A central challenge is \emph{cross-stage coupling}: choices made in earlier stages can affect the optimization regime of downstream stages, so optimizing one stage in isolation does not reliably optimize the full pipeline. For instance,~\citet{kang2025quagmires} found that improvements in SFT metrics do not necessarily translate into better post-RL performance. As a result, expert-driven tuning often becomes an expensive trial-and-error process that can still yield suboptimal end-to-end performance. Yet automation is also difficult under realistic compute budgets.

Recent approaches reduce this problem to an online black-box hyperparameter optimization (HPO) problem, in which configurations are sequentially proposed and evaluated through expensive end-to-end pipeline executions, so only a small number of trials are feasible in practice. Without a strong prior, such online search also suffers from cold-start inefficiency, wasting early trials in low-quality regions of the search space. This cost can in principle be reduced through multi-fidelity techniques~\citep{li2017hyperband}. However, in multi-stage post-training with RL, partial-training signals can be noisy and only weakly, or even non-monotonically, related to the final post-RL outcome, making them unreliable when used in isolation as direct optimization signals. Purely offline solutions are also insufficient: users often optimize on proprietary or domain-specific datasets whose distributions may not appear in historical runs, and offline surrogates learned from past data can suffer from dataset shift. 

These challenges motivate a hybrid approach that (i) leverages offline experience to provide a transferable inductive bias over promising configurations, (ii) adapts online to the target dataset using a small number of evaluations, and (iii) uses proxy signals from truncated runs to reduce the cost of online adaptation.

We propose \emph{AutoPipe}, a two-phase framework for budgeted configuration selection in coupled SFT-RL LLM post-training pipelines. In the {offline} phase, AutoPipe learns a {dataset-conditioned learning-to-rank} surrogate from historical post-training runs, capturing within-dataset preferences among configurations and providing transferable guidance toward promising regions of the search space. In the {online} phase, for a new target dataset, AutoPipe uses these offline ranking scores as meta-learned guidance and performs Bayesian optimization over residual corrections, concentrating limited online capacity on correcting dataset-specific miscalibration. To reduce evaluation cost, each online candidate is assessed via a truncated (early-stopped) run; a learned {early-stop predictor} maps early training signals to a calibrated proxy of final performance, which is then used to support robust online updates under noisy feedback. Across biomedical question answering tasks, AutoPipe achieves performance comparable to full-fidelity baselines while using a fraction of their online cost, and outperforms lower-cost early-stopped baselines at a modestly higher compute budget.

\paragraph{Contributions.} (1) We propose \emph{AutoPipe}, a budget-aware framework for configuration selection in SFT-RL post-training pipelines. (2) We develop a dataset-conditioned ranking surrogate that transfers configuration preferences across datasets and offers strong offline guidance toward promising candidates. (3) We introduce an online adaptation mechanism based on Gaussian-process residual modeling and an early-stop proxy predictor to correct dataset-specific miscalibration under tight evaluation budgets. (4) Extensive experiments on biomedical reasoning datasets show that AutoPipe consistently outperforms strong offline-only baselines and remains competitive with full-fidelity online baselines at substantially lower online adaptation cost.

\section{Related Work}

\paragraph{Hyperparameter Optimization.}
 Bayesian optimization (BO) is a standard approach for sample-efficient hyperparameter optimization \citep{snoek2012practical}. Systems such as SMAC \citep{hutter2011sequential} and TPE \citep{bergstra2011algorithms} extend BO to mixed discrete-continuous spaces, while multi-fidelity methods like Hyperband \citep{li2017hyperband} and BOHB \citep{falkner2018bohb} adaptively allocate compute. These methods often assume low-fidelity signals predict final performance; in multi-stage LLM post-training, early SFT may correlate weakly with downstream RL outcomes. Classic BO can also require many evaluations, which is impractical when full pipelines are expensive.

\paragraph{Meta- and Transfer-Learning.}
To reduce optimization costs, we may leverage traces from related tasks via similarity-weighted surrogate ensembles \citep{feurer2015efficient}, two-phase offline-to-online procedures \citep{wistuba2016two, wu2026reward}, or learned priors and search-space biases from meta-datasets \citep{perrone2018scalable}. Recent offline recommendation solutions such as Quick-Tune \citep{arango2024quick} often treat learned priors or predictors as static initializations and can fail under distribution shift. This motivates hybrid strategies that reuse offline signal while adapting to new observations.

\paragraph{Automatic LLM Post-Training Pipelines.}
The modern paradigm combining SFT with RL was popularized by InstructGPT \citep{ouyang2022training}. Configuration choices across stages substantially affect final quality, yet end-to-end optimization remains limited. Recent work optimizes components of this process, including black-box search for instruction-tuning hyperparameters \citep{tribes2024hyperparameter}, multi-objective optimization of LLM and RAG pipelines at inference time \citep{barker2025faster}, and agent-based frameworks for pipeline construction such as LaMDAgent \citep{yano2025lamdagent}. In contrast, our work targets end-to-end SFT+RL pipelines, emphasizing budget-limited optimization by leveraging historical runs while allowing online adaptation to new datasets.

\section{Methodology}
\label{sec:method}

\begin{figure}[t]
\centering
\includegraphics[width=\columnwidth]{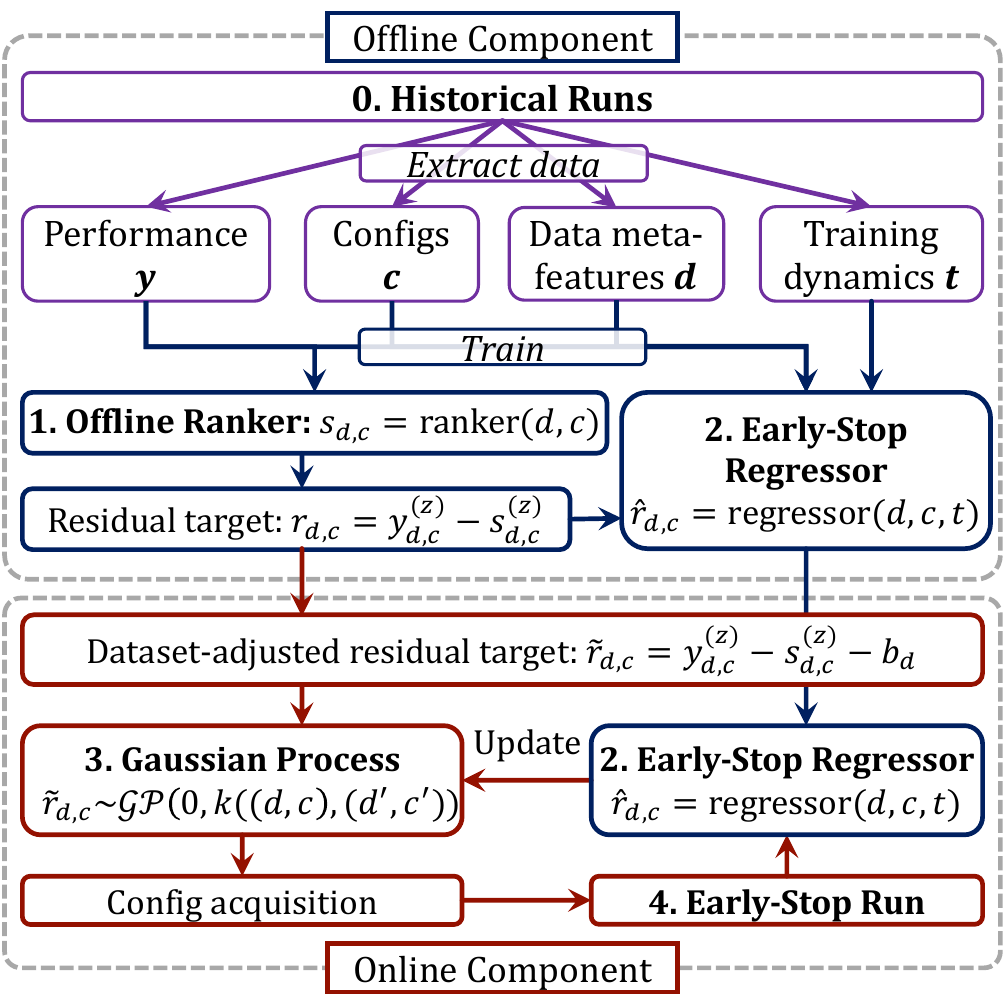} 
\caption{Overview of the proposed two-phase configuration selection framework.}
\label{fig:method-overview}
\end{figure}

Modern LLM applications are increasingly implemented as \emph{multi-stage} systems, commonly involving components such as SFT, retrieval augmentation, memory \citep{wu2026towards}, RL, and routing mechanisms; this motivates system-level optimization that accounts for cross-component dependencies \cite{wadlom2026efficient}. In this work, we focus on a widely deployed core pipeline consisting of SFT followed by RL.  

Given a target dataset, post-training performance depends on a \emph{large, heterogeneous} configuration space spanning both discrete and continuous choices (e.g., base model, batch sizes and learning rate schedule). End-to-end evaluation of a single configuration requires running the full pipeline, including expensive RL rollouts, thus prohibiting exhaustive or even moderately broad search. This motivates methods that (i) leverage historical evidence across datasets and (ii) allocate limited online compute toward only the most promising candidates.

\vspace{0.05in}\noindent\textbf{Problem Formulation}. 
Let $d$ denote a dataset and $c \in \mathcal{C}$ denote a post-training configuration covering SFT and RL hyperparameters. Running the pipeline with $(d,c)$ yields a performance score $y(d,c)$ (e.g., accuracy on a target evaluation suite). Given a new dataset $d^\star$, the goal is to identify the best configuration
\begin{equation}
c^\star \in \arg\max_{c \in \mathcal{C}} y(d^\star, c),
\end{equation}
subject to a strict online evaluation budget measured in number of evaluations.

Two considerations guide our design. First, our objective is selection: for each dataset, we aim to identify the best configuration (or a small set of top candidates), rather than precisely predict absolute performance. Second, datasets differ in ways that break global consistency, that is, the relative ordering of configurations is not stable across datasets. Hence, we seek to learn \emph{dataset-conditioned} preferences over configurations. 

\vspace{0.05in}\noindent\textbf{Two-Phase Optimization}. We describe our solution below and in Figure~\ref{fig:method-overview}:
\begin{enumerate}[leftmargin=*,nolistsep]
    \item \emph{Offline phase (amortized learning):} Using historical runs across datasets, we learn a dataset-conditioned {ranking surrogate} that produces configuration ranking scores $s_{d,c}$.
    \item \emph{Online phase (budgeted adaptation):} For a new dataset $d^\star$, the surrogate scores $s_{d^\star,c}$ are used as {meta-learned guidance} within BO. To reduce evaluation cost, configurations are executed via {early-stopped} runs whose outcomes are mapped to proxy scores by an early-stop predictor.
\end{enumerate}

\noindent This design separates {knowledge acquisition} (offline) from {dataset-specific adaptation} (online), allowing for historical evidence to be exploited while remaining flexible enough to correct for dataset-specific deviations under tight compute constraints.
Below we elaborate on these two phases. 

\vspace{0.05in}\noindent\textbf{Offline Surrogate Learning}. We assume an offline corpus of completed runs, each consisting of a dataset $d$, configuration $c$, and performance score $y(d,c)$ (e.g., average benchmark accuracy). Runs are organized into \emph{dataset groups}, where each group contains evaluations of multiple configurations on the same dataset.

\vspace{0.05in}\noindent\emph{Ranking supervision.} Rather than predicting absolute performance, we formulate the surrogate learning problem as \emph{learning to rank}. For each dataset group, observed outcomes induce a partial ordering over configurations. Training supervision thus consists of pairwise preference constraints of the form $c_i \succ c_j$ whenever $y(d,c_i) > y(d,c_j)$. We restrict comparisons to \emph{within-dataset} comparisons to reduce sensitivity to cross-dataset shifts in performance scale, focusing the surrogate on identifying the most promising configurations for a given dataset.

\vspace{0.05in}\noindent\emph{Surrogate model.} We instantiate the surrogate using a gradient-boosted tree ranker (XGBRanker) trained with the pairwise ranking objective (\texttt{rank:pairwise}). Each example uses features
\[
\phi(d,c) = \big[\phi_{\text{cfg}}(c)\ ;\ \phi_{\text{meta}}(d)\big],
\]
where $\phi_{\text{cfg}}$ encodes SFT/RL hyperparameters and $\phi_{\text{meta}}$ encodes dataset features. During inference, the surrogate scores candidate configurations for a target dataset $d^\star$, producing ranking scores $s_{d^\star,c}$ that estimate the relative quality of configurations.
We adopt gradient-boosted trees for their robustness in low-data regimes and their ability to capture non-linear interactions between configuration choices and dataset properties. Detailed feature definitions, preprocessing steps, and surrogate hyperparameters are discussed in \cref{appendix:offline}.

\vspace{0.05in}\noindent\textbf{Online BO with Meta-Learned Guidance}. 
The offline ranker captures transferable structure in configuration quality, but directly selecting configurations by $s_{d^\star,c}$ can be suboptimal due to shifts in performance landscape for the new dataset $d^\star$. We thus perform BO using the offline score as meta-learned guidance, adapting to the target dataset through a small number of online evaluations.

\vspace{0.05in}\noindent\emph{Offline-Guided Decomposition.}
Let $s_{d^\star,c}^{(z)}$ denote the standardized offline score assigned by the ranker to configuration $c$ for dataset $d^\star$. In the online phase, we do not model absolute performance from scratch; instead, we maintain a GP over dataset-specific correction terms relative to this offline guidance. Conceptually, we define the correction target as
\begin{equation}
\tilde{r}_{d^\star,c} = y_{d^\star,c}^{(z)} - s_{d^\star,c}^{(z)} - b_{d^\star},
\end{equation}
where $b_{d^\star}$ is a dataset-specific offset that captures systematic mismatch between the offline score and the target dataset. The GP models configuration-dependent variation in this correction term:
\begin{equation}
f_{d^\star}(c) \approx \tilde{r}_{d^\star,c}, \quad
f_{d^\star}(c) \sim \mathcal{GP}(0, k(c,c')).
\end{equation}
In practice, the online procedure never requires direct access to true full-fidelity values $y_{d^\star,c}^{(z)}$ during search. Instead, it updates the GP using low-cost pseudo-observations derived from the early-stop predictor, while the offline score $s_{d^\star,c}^{(z)}$ serves as a persistent guidance signal throughout optimization, with its influence modulated over iterations. For the GP, we use a Mat\'ern-$5/2$ kernel with automatic relevance determination (ARD) over configuration dimensions, providing robustness against non-uniform sensitivity in the configuration space~\cite{snoek2012practical}.

\vspace{0.05in}\noindent\emph{Posterior prediction and tempered guidance.}
At BO iteration $t$, let $\mu_{f,t}(c)$ and $\sigma_{f,t}(c)$ denote the GP posterior mean and standard deviation over the correction term. We form the acquisition score using a tempered combination of offline guidance and online correction:
\begin{equation}
\hat{y}_{d^\star,c,t}^{(z)} = w_t\, s_{d^\star,c}^{(z)} + b_{d^\star,t} + \mu_{f,t}(c),
\end{equation}
where $b_{d^\star,t}$ is the current dataset-specific offset estimate and $w_t \in [0,1]$ controls how strongly the search remains anchored to the offline ranker. We linearly anneal $w_t$ over iterations so that early search is guided more strongly by transferable cross-dataset structure, while later iterations place greater weight on dataset-specific evidence.

\vspace{0.05in}\noindent\emph{Acquisition and posterior updates.}
We select the next configuration by maximizing an Upper Confidence Bound objective,
\begin{equation}
\alpha_t(c) = \hat{y}_{d^\star,c,t}^{(z)} + \beta_t\, \sigma_{f,t}(c),
\end{equation}
where $\beta_t$ controls the exploration--exploitation trade-off. Since $s_{d^\star,c}^{(z)}$ is fixed across iterations, exploration is driven by uncertainty in the residual correction, while the overall search trajectory remains anchored to the meta-learned guidance. Each selected configuration is then evaluated via the early-stop predictor; the resulting pre-bias residual pseudo-observation is incorporated into the online update by first refreshing the dataset-specific offset and then refitting the GP on centered targets. Additional implementation details are provided in \cref{appendix:online}.

\vspace{0.05in}\noindent\textbf{Early-Stop Predictor}. 
Full end-to-end SFT+RL evaluation is too expensive to serve as the online feedback signal. We instead employ an \emph{early-stop predictor} that uses partial training signals from an inexpensive truncated run to approximate the full evaluation outcome. The key idea is to exploit the fact that downstream RL performance depends strongly on the quality and stability of the SFT initialization, making early SFT dynamics a useful source of low-cost evidence for online updates.

\vspace{0.05in}\noindent\emph{Predicting standardized residual corrections.}
For each executed run, we extract early training features such as summary statistics computed from logged training dynamics, and combine with configuration and dataset features to form $x_{\mathrm{es}}(d,c)$. Rather than regressing directly on absolute end-to-end performance, we train the predictor to estimate the standardized residual correction
\begin{equation}
r_{d,c} = y_{d,c}^{(z)} - s_{d,c}^{(z)},
\end{equation}
where $y_{d,c}^{(z)}$ is the standardized full evaluation score and $s_{d,c}^{(z)}$ is the standardized offline ranking score. We instantiate $g(\cdot)$ as an ensemble of XGBoost regressors (XGBRegressor), trained to predict
\begin{equation}
\hat{r}_{d,c} = g(x_{\mathrm{es}}(d,c)) \approx r_{d,c},
\end{equation}
so that the final proxy score is
\begin{equation}
\hat{y}_{d,c}^{(z)} = s_{d,c}^{(z)} + \hat{r}_{d,c}.
\end{equation}
This makes the predictor a conservative correction mechanism: when early-stop signals are uninformative, predictions naturally remain anchored to the offline score.

\vspace{0.05in}\noindent\emph{Use in online optimization.}
For a new dataset $d^\star$, each candidate is run only to the early-stop point, from which $x_{\mathrm{es}}(d^\star,c)$ is computed. The predictor then produces a residual correction $\hat{r}_{d^\star,c}$, which is used as a pseudo-observation to update the GP in residual space. The reconstructed score
\begin{equation}
\hat{y}_{d^\star,c}^{(z)} = s_{d^\star,c}^{(z)} + \hat{r}_{d^\star,c}
\end{equation}
serves as a low-cost proxy for comparing candidate configurations and deciding which ones merit full end-to-end evaluation. Feature definitions, training and other implementation details are deferred to \cref{appendix:early_stop_regressor}.

\vspace{0.05in}\noindent\textbf{Remark}. Our key design principle is to separate \emph{cross-dataset regularities} from \emph{dataset-specific correction}. The offline stage amortizes knowledge about configuration quality across datasets and supplies meta-learned guidance, while the online stage refines it through residual updates from inexpensive early-stopped runs. This enables adaptive configuration selection under tight evaluation budgets.

\section{Experiments}
\label{sec:exp}

\subsection{Experimental Setup}
\label{sec:exp:protocol}
We evaluate on biomedical chain-of-thought question answering (QA), a setting with datasets varying in style and reasoning depth, thereby providing a meaningful testbed for configuration transfer.

\begin{table}[h]
\small
\centering
\caption{Post-training datasets in our LODO evaluation.}
\label{tab:datasets}
\begin{tabular}{clc}
\toprule
\textbf{\#} & \textbf{Dataset} & \textbf{Reference} \\
\midrule
1 & MQA & Ours \\
2 & MQAR & Ours \\
3 & PQA & Ours \\
4 & MQA(R) & Ours \\
5 & MQA(R)PQA & Ours \\
6 & HuaTuo & \citep{chen2024huatuogpt} \\
7 & MedReason & \citep{wu2025medreason} \\
8 & MedS3 & \citep{jiang2025meds} \\
9 & ReasonMed & \citep{sun2025reasonmed} \\
10 & m23k & \citep{huang2026m1} \\
\bottomrule
\end{tabular}
\end{table}

\begin{table}[h]
\centering
\small
\caption{Downstream benchmarks in our evaluation.}
\label{tab:benchmarks}
\begin{tabularx}{\columnwidth}{lX}
\toprule
\textbf{Benchmark} & \textbf{Reference} \\
\midrule
BioASQ (yes/no) & \citep{xiong2024benchmarking} \\
GPQA (Biology) & \citep{rein2024gpqa} \\
MedBullets (4 options) & \citep{chen2025benchmarking} \\
MedBullets (5 options) & \citep{chen2025benchmarking} \\
MedMCQA & \citep{pal2022medmcqa} \\
MedQA & \citep{jin2021disease} \\
MedXpertQA & \citep{zuo2025medxpertqa} \\
MMLU (Medicine) & \citep{hendrycks2021measuring} \\
MMLU-Pro (Bio \& Health) & \citep{wang2024mmlu} \\
PubMedQA & \citep{jin2019pubmedqa} \\
PubMedQA (questions only) & \citep{jin2019pubmedqa} \\
\makecell[tl]{PubMedQA (with \\ \hspace{2ex} distractor abstracts)} & \citep{jin2019pubmedqa}, this work \\
\bottomrule
\end{tabularx}
\end{table}

\vspace{0.05in}\noindent\textbf{Datasets and Benchmarks.}
We use a diverse collection of medical reasoning datasets for post-training and evaluate selected configurations on a fixed suite of downstream biomedical benchmarks. Half of the datasets are newly constructed in this work, while the remainder are open-source datasets from recent medical reasoning papers. Tables~\ref{tab:datasets} and~\ref{tab:benchmarks} summarize datasets and benchmarks; detailed descriptions are provided in \cref{appendix:dataset}. 

\emph{Transfer protocol.} We evaluate transfer to unseen datasets using a leave-one-dataset-out (LODO) protocol. In each split, one dataset is held out as the target $d^\star$, and the remaining datasets form the offline corpus used to train both the ranking surrogate and early-stop regressor. The online procedure is then run on $d^\star$ under a fixed number of online evaluations, reflecting a practical deployment setting in which extensive prior experimentation is infeasible. Stochastic methods are repeated across 3 random seeds. Additional details are provided in \cref{appendix:e2eeval}.

\emph{Source-disjoint evaluation.} Some LODO folds share underlying question sources. In particular, MQA(R) and MQA(R)PQA are constructed by sampling from MQA/MQAR and MQA/MQAR/PQA respectively, and certain common sources such as MedQA \cite{jin2021disease} also occur across multiple folds. To evaluate transfer under a stronger distribution shift, we additionally consider a setting in which all ten datasets in Table~\ref{tab:datasets} serve as historical data and the unseen target $d^\star$ is the source-disjoint CasiMedicos-Arg \cite{sviridova-etal-2024-casimedicos}, whose question source is absent from all other ten datasets.

\vspace{0.05in}\noindent\textbf{Configuration space.}
All methods search a shared configuration space spanning base model choice and SFT/RL hyperparameters.
Table~\ref{tab:configspace} summarizes the space; \cref{appendix:configspace} details how we sample from the space to construct the offline corpus.

\begin{table}[t!]
\small
\centering
\caption{Configuration space searched by all methods.}
\label{tab:configspace}
\begin{tabular}{p{0.6cm} p{2.5cm} l}
\toprule
\textbf{Phase} & \textbf{Hyperparameter} & \textbf{Possible values} \\
\midrule
& \multirow{6}{*}{Base model} & Qwen2.5-1.5B-Instruct \\
& & Qwen2.5-3B-Instruct \\
& & Qwen2.5-7B-Instruct \\
& & Llama3.2-1B-Instruct \\
& & Llama3.2-3B-Instruct \\
& & Llama3.1-8B-Instruct \\
\midrule
\multirow{3}{*}{SFT} & Number of epochs & 1, 2, 3 \\
& Effective batch size & 32, 64, 128 \\
& Learning rate & $1e^{-5}, 2e^{-5}, 5e^{-5}$ \\
\midrule
\multirow{5}{*}{RL} & Effective batch size & 64, 128, 256 \\
& Learning rate & $1e^{-6},2e^{-6},5e^{-6},1e^{-5}$ \\
& Beta & 0, 0.05, 0.1 \\
& Number of rollouts & 8, 16 \\
& Temperature & 0.7, 0.9, 1.1 \\
\bottomrule
\end{tabular}
\end{table}

\vspace{0.05in}\noindent\textbf{Baselines.}
We compare against two types of baselines: (1) \emph{static / zero-shot} methods, which do not perform any online evaluations on the target dataset, and (2) \emph{sequential} methods, which adapt through online evaluations under the budget constraint. Under \emph{static / zero-shot baselines}, we consider: (1) \emph{Random}: one random configuration; (2) \emph{LLM Recommendation (Description-only)}: prompts an LLM to propose a configuration, given only task and dataset description; (3) \emph{LLM Recommendation (History-conditioned)}: additionally provides a set of similar historical datasets with evaluated configurations and outcomes; (4) \emph{Global}: selects the configuration that achieved the best performance across the offline corpus (excluding the held-out dataset); (5) \emph{Offline-only}: selects $\arg\max_c (s_{d^\star,c})$ without online adaptation. For \emph{sequential baselines}, we compare: (1) \emph{Random Search}: randomly samples configurations; (2) \emph{BOHB} \citep{falkner2018bohb}: combines multi-fidelity resource allocation with kernel density estimators to prioritize promising configurations; (3) \emph{SMAC} \citep{hutter2011sequential}: uses random-forest-based BO for algorithm configuration; (4) \emph{TPE} \citep{bergstra2011algorithms}: fits separate density models to high- and low-performing regions of the search space and selects configurations that maximize their density ratio; (5) \emph{RGPE} \cite{feurer2018scalable}: constructs an ensemble of surrogate models learned on previous tasks and learns to reweight them based on target task relevance; (6) \emph{AutoPipe}, our full solution. Baseline implementation details are provided in \cref{appendix:baseline}.

\vspace{0.05in}\noindent\textbf{Metrics}. We report metrics that capture end-to-end performance and selection quality under a limited number of online evaluations.

\emph{Pipeline accuracy.}
We report the (1) \emph{average benchmark score} achieved by the configuration $\hat{c}$ selected by each method after full post-training, (2) the corresponding \emph{regret} that quantifies the optimality gap, $y(d^\star, c^\star_{\mathrm{opt}}) - y(d^\star, \hat{c})$, where $c^\star_{\mathrm{opt}}$ is the best-performing configuration in the reference pool for $d^\star$ (i.e., the set of all configurations evaluated on $d^\star$ anywhere in this study, rather than the entire search space) and (3) the \emph{evaluation cost}, measured in H200 GPU hours, which includes both online adaptation cost and the cost of evaluating the selected configuration. This cost excludes construction of the historical corpus, which is a one-time unpriced cost that is reused across target datasets and therefore amortized rather than charged to each adaptation run. 

\emph{Ranking and selection diagnostics.} For component-level evaluation, we additionally report diagnostics for the models trained on the offline corpus, namely the offline surrogate and the early-stop predictor. Specifically, we report: (1) \emph{Recall@$k$}, whether $c^\star_{\mathrm{opt}}$ appears in the top-$k$ ranked configurations; (2) \emph{nDCG@$k$}, which emphasizes correct ordering near the top of the ranking; (3) \emph{Pairwise accuracy}, the fraction of configuration pairs ordered correctly within each dataset; and (4) \emph{Spearman correlation}, the rank correlation between predicted and ground-truth within-dataset orderings. 

\begin{table}[t!]
\centering
\small
\caption{LODO configuration selection performance under a fixed number of online evaluations using either full-fidelity signals or early-stop (ES) proxy signals. For stochastic methods, results are reported as mean \(\pm\) standard error over three independent repetitions; see Appendix C.1 for details.} 
\label{tab:e2e-proxy}
\resizebox{\columnwidth}{!}{%
\begin{tabular}{lccc}
\toprule
\textbf{Method} & \textbf{Performance} & \textbf{Regret} & \textbf{Cost} \\
\midrule
\rowcolor[gray]{.95} \multicolumn{4}{l}{\textit{Static \& Zero-Shot Baselines}} \\
Random & 0.362 $\pm$ 0.015 & 0.289 $\pm$ 0.015 & 4.5 $\pm$ 0.4 \\
LLM (Desc.) & 0.307 & 0.343 & 7.6 \\
LLM (Hist.) & 0.442 & 0.208 & 5.6 \\
Global & 0.529 & 0.122 & 4.1 \\
Offline-only & 0.557 $\pm$ 0.016 & 0.094 $\pm$ 0.016 & 6.4 $\pm$ 0.2 \\
\midrule
\rowcolor[gray]{.95} \multicolumn{4}{l}{\textit{Sequential Baselines}} \\
\multicolumn{2}{l}{Random Search} & & \\
\hspace{3mm} ES & 0.510 $\pm$ 0.033 & 0.140 $\pm$ 0.033 & 6.2 $\pm$ 0.6 \\
\hspace{3mm} Full & 0.573 $\pm$ 0.007 & 0.077 $\pm$ 0.007 & 61.7 $\pm$ 1.7 \\
BOHB & 0.593 $\pm$ 0.002 & 0.057 $\pm$ 0.002 & 98.5 $\pm$ 2.1 \\
RGPE & & & \\
\hspace{3mm} ES & 0.534 $\pm$ 0.013 & 0.116 $\pm$ 0.013 & 9 $\pm$ 0.2 \\
\hspace{3mm} Full & 0.599 $\pm$ 0.002 & 0.052 $\pm$ 0.002 & 75.6 $\pm$ 2.1 \\
SMAC & & & \\
\hspace{3mm} ES & 0.511 $\pm$ 0.003 & 0.139 $\pm$ 0.003 & 6.0 $\pm$ 0.1 \\
\hspace{3mm} Full & 0.590 $\pm$ 0.006 & 0.060 $\pm$ 0.006 & 65.7 $\pm$ 1.1 \\
TPE & & & \\
\hspace{3mm} ES & 0.539 $\pm$ 0.006 & 0.111 $\pm$ 0.006 & 6.0 $\pm$ 0.2 \\
\hspace{3mm} Full & 0.588 $\pm$ 0.010 & 0.062 $\pm$ 0.010 & 65.2 $\pm$ 0.8 \\
\midrule
Ours & 0.594 $\pm$ 0.002 & 0.056 $\pm$ 0.002 & 9.5 $\pm$ 0.4 \\
\bottomrule
\end{tabular}
}
\end{table}

\subsection{Results and Analysis}
\label{sec:results}

We evaluate the framework to assess overall end-to-end effectiveness and to clarify the contribution of its components. Our analysis has four parts: (i) configuration selection under low-cost proxy-feedback and higher-cost full-fidelity optimization regimes, measured by full-fidelity performance of the selected configuration; (ii) transfer to a source-disjoint target whose question source is entirely distinct from those in the historical corpus; (iii) ablations of key design choices; and (iv) sensitivity to online budget, early-stop fidelity and online optimization parameters.

\vspace{0.05in}\noindent\textbf{Efficacy across budget regimes}. 
We study configuration selection when optimization uses early-stopped, noisy evaluations, but success is measured by full end-to-end performance. We compare proxy-guided methods in a resource-constrained regime against sequential optimizers that operate on full-fidelity evaluations. In the constrained regime, methods use early-stopped runs scored by the early-stop predictor; after the budget is exhausted, the selected configuration is verified with a full-fidelity run.

Table~\ref{tab:e2e-proxy} reports full-fidelity performance, regret and optimization cost across offline and online baselines. The results shed light on the role of {learned offline structure} and {online adaptation} in configuration selection. Firstly, the offline ranker is already a strong baseline, establishing a performance floor that outperforms other static baselines. While random selection and LLM-based recommendations struggle with the interaction-heavy post-training landscape, the ranker captures transferable patterns across datasets, producing a markedly improved initial configuration ordering. The performance of the experience-informed yet dataset-agnostic \emph{Global} baseline further highlights the importance of dataset conditioning. Although it achieves competitive average performance, it exhibits higher variability across datasets, performing comparably or even better than dataset-conditioned selection in certain cases while incurring substantial failures in others (dataset-level comparisons are provided in Appendix~\ref{appendix:results-e2e}). This suggests that experience alone is insufficient; without dataset conditioning, strong average performance can mask brittle behavior under distributional shift.

Secondly, incorporating the online phase further elevates performance, bringing the full two-phase solution into a regime competitive with expensive sequential baselines like BOHB, RGPE, SMAC, and TPE. Notably, paired two-sided Wilcoxon signed-rank tests on per-dataset mean performance showed that AutoPipe significantly outperforms the Offline-only baseline (Holm-adjusted $p = 0.027$), while its difference from the highest-performing full-fidelity baseline, RGPE, is not statistically significant ($p = 0.287$). Full-fidelity methods, which make use of multiple end-to-end evaluation cycles, represent a high-resource point of comparison. Their early-stopped variants improve over several weaker static baselines but remain below their full-fidelity counterparts, indicating that early-stop signals are noisy yet directionally informative. This gap underscores the value of offline structure: proxy signals alone are insufficient, but when used to refine a strong dataset-conditioned prior, they can recover much of the benefit of sequential search.

Overall, the end-to-end results show that competitive configuration selection is achievable even when online optimization is limited to early-stopped signals, provided that these signals are combined with structure learned across datasets. The two-phase framework narrows the efficiency-quality gap by amortizing global interaction structure offline and reserving online computation for dataset-specific correction. 

\begin{table}[t]
\centering
\small
\caption{LODO ranking performance of offline surrogate and early-stop (ES) regressor.}
\label{tab:offline-ranking}
\begin{tabular}{lcc}
\toprule
\textbf{Metric} & \textbf{Offline Ranker} & \textbf{ES Regressor} \\
\midrule
Regret & 0.118 $\pm$ 0.013 & 0.096 $\pm$ 0.017 \\
Recall@1 & 0.100 $\pm$ 0.000 & 0.100 $\pm$ 0.000 \\
Recall@3 & 0.300 $\pm$ 0.058 & 0.233 $\pm$ 0.067 \\
Recall@5 & 0.400 $\pm$ 0.058 & 0.300 $\pm$ 0.000 \\
Recall@10 & 0.633 $\pm$ 0.033 & 0.467 $\pm$ 0.033 \\
nDCG@5 & 0.840 $\pm$ 0.007 & 0.851 $\pm$ 0.005 \\
nDCG@10 & 0.854 $\pm$ 0.009 & 0.856 $\pm$ 0.006 \\
Pairwise Accuracy & 0.747 $\pm$ 0.002 & 0.764 $\pm$ 0.000 \\
Spearman $\rho$ & 0.675 $\pm$ 0.005 & 0.707 $\pm$ 0.001 \\
\bottomrule
\end{tabular}
\end{table}

\vspace{0.05in}\noindent\textbf{Offline Component Analysis}.
We next examine whether the offline components provide useful signals before online adaptation. Table~\ref{tab:offline-ranking} reports ranking metrics between true end-to-end performance and predictions for the offline ranker and early-stop regressor. The strong performance of the offline ranker suggests that it transfers useful ordering structure across datasets despite heterogeneous performance scales. The early-stop regressor is competitive with, and on several metrics slightly stronger than, the offline ranker, while serving a different role: it predicts dataset-specific residual corrections from partial training dynamics, thus acting as a cautious correction mechanism rather than a standalone selector. An analysis of the feature groups used by the offline ranker and early-stop regressor is included in Appendix~\ref{appendix:featsanalysis}; broadly, dataset-derived features account for most of the normalized gain in both models, while training dynamics features also play a substantive role for the early-stop regressor.

\subsection{Source-Disjoint Transfer}
\label{sec:source-disjoint-transfer}

The primary LODO setting contains source relationships among some of the post-training datasets, particularly for sources such as MedQA \cite{jin2021disease} and PubMedQA \cite{jin2019pubmedqa}. We therefore evaluate AutoPipe on CasiMedicos-Arg \cite{sviridova-etal-2024-casimedicos}, whose question source of Spanish medical exams is absent from the other ten datasets used in the LODO setting. This represents a stronger transfer setting in which offline components cannot benefit from historical runs with overlapping question sources. Table~\ref{tab:casicomedicos} reports the results. Similarly to our main LODO evaluation, AutoPipe achieves results comparable to its full-fidelity counterparts, demonstrating that combining historical guidance with target-specific online correction confers benefits even when the target question source is disjoint from those in the offline corpus. Component ablations and sensitivity analyses on CasiMedicos-Arg show broadly similar qualitative behavior to the primary LODO experiments and are reported in Appendix~\ref{appendix:casimedicos}.

\begin{table}[t!]
\centering
\small
\caption{Configuration selection performance, regret and cost on the source-disjoint dataset.} 
\label{tab:casicomedicos}
\resizebox{\columnwidth}{!}{%
\begin{tabular}{lccc}
\toprule
\textbf{Method} & \textbf{Performance} & \textbf{Regret} & \textbf{Cost} \\
\midrule
\rowcolor[gray]{.95} \multicolumn{4}{l}{\textit{Static \& Zero-Shot Baselines}} \\
Random & 0.452 $\pm$ 0.068 & 0.175 $\pm$ 0.068 & 0.3 $\pm$ 0.1 \\
LLM (Desc.) & 0.019 & 0.608 & 0.4 \\
LLM (Hist.) & 0.518 & 0.109 & 0.3 \\
Global & 0.181 & 0.446 & 0.7 \\
Offline-only & 0.592 $\pm$ 0.003 & 0.035 $\pm$ 0.003 & 0.5 $\pm$ 0.1 \\
\midrule
\rowcolor[gray]{.95} \multicolumn{4}{l}{\textit{Sequential Baselines}} \\
\multicolumn{2}{l}{Random Search} & & \\
\hspace{3mm} ES & 0.552 $\pm$ 0.029 & 0.075 $\pm$ 0.029 & 0.6 $\pm$ 0.1 \\
\hspace{3mm} Full & 0.595 $\pm$ 0.005 & 0.032 $\pm$ 0.005 & 7.3 $\pm$ 0.1 \\
BOHB & 0.600 $\pm$ 0.005 & 0.027 $\pm$ 0.005 & 9.0 $\pm$ 0.5 \\
RGPE & & & \\
\hspace{3mm} ES & 0.614 $\pm$ 0.003 & 0.013 $\pm$ 0.003 & 1.0 $\pm$ 0.1 \\
\hspace{3mm} Full & 0.622 $\pm$ 0.003 & 0.005 $\pm$ 0.003 & 9.7 $\pm$ 1.0 \\
SMAC & & & \\
\hspace{3mm} ES & 0.549 $\pm$ 0.032 & 0.078 $\pm$ 0.032 & 0.8 $\pm$ 0.1 \\
\hspace{3mm} Full & 0.618 $\pm$ 0.006 & 0.009 $\pm$ 0.006 & 6.7 $\pm$ 0.5 \\
TPE & & & \\
\hspace{3mm} ES & 0.530 $\pm$ 0.007 & 0.097 $\pm$ 0.007 & 0.6 $\pm$ 0.1 \\
\hspace{3mm} Full & 0.590 $\pm$ 0.012 & 0.037 $\pm$ 0.012 & 6.8 $\pm$ 1.0 \\
\midrule
Ours & 0.624 $\pm$ 0.002 & 0.003 $\pm$ 0.002 & 1.1 $\pm$ 0.1 \\
\bottomrule
\end{tabular}
}
\end{table}

\subsection{Ablation Studies}
\label{sec:results-ablation}

We ablate three key components of the framework to verify that observed gains arise from their interaction rather than from any single design choice: (1) the residual formulation of the GP, (2) the BO procedure itself, and (3) the offline ranker.

\vspace{0.05in}\noindent\emph{No Residual.} This ablation removes the residual formulation, where the GP models corrections to the ranking score ($y_{d,c}^{(z)}-s_{d,c}^{(z)}-b_d$), and instead models $y_{d,c}^{(z)}$ directly. The offline ranker is still used to guide selection. This tests whether learning residuals from a strong prior improves sample efficiency compared to learning the objective from scratch.

\vspace{0.05in}\noindent\emph{No BO.} This ablation removes the BO loop and evaluates the top $k$ configurations ranked by the offline ranker. This tests whether adaptive search adds value beyond trusting the offline ranker alone.

\vspace{0.05in}\noindent\emph{No Offline.} This ablation removes all involvement of the offline ranker from the online phase. The GP models $y_{d,c}^{(z)}$ directly, candidate selection is no longer guided by the ranker, and the early-stop regressor is retrained to predict $y_{d,c}^{(z)}$ rather than residuals. This tests the overall value of incorporating offline knowledge into the search process.

\begin{table}[h]
\centering
\small
\setlength{\tabcolsep}{4pt}
\caption{Ablation results for key components.}
\label{tab:ablation}
\begin{tabular}{lccc}
\toprule
\textbf{Variant} & \textbf{Performance} & \textbf{Regret} & \textbf{Cost} \\
\midrule
Full method & 0.594 $\pm$ 0.002 & 0.056 $\pm$ 0.002 & 9.5 $\pm$ 0.4 \\
\midrule
No Residual & 0.549 $\pm$ 0.017 & 0.101 $\pm$ 0.017 & 9.0 $\pm$ 0.5 \\
No BO & 0.552 $\pm$ 0.019 & 0.098 $\pm$ 0.019 & 9.9 $\pm$ 0.4 \\
No Offline & 0.511 $\pm$ 0.020 & 0.139 $\pm$ 0.020 & 7.1 $\pm$ 0.8 \\
\bottomrule
\end{tabular}
\end{table}

\noindent Table~\ref{tab:ablation} clarifies the role of each component in the framework. Removing any single one degrades performance, with the largest drop occurring when offline structure is removed entirely in the \emph{No Offline} variant. This variant exhibits performance comparable to the early-stopped versions of the sequential baselines, suggesting that without offline learned guidance, the method largely reverts to classical BO under noisy and weakly informative signals. 
The \emph{No Residual} ablation indicates that directly modeling absolute performance instead of residual corrections leads to a substantial performance drop, suggesting that correcting the offline score is more sample-efficient than learning the full objective from noisy proxy observations.  
Additionally, \emph{No BO} slightly outperforms \emph{No Residual} despite the removal of adaptive search, suggesting that a structured prior with conservative correction can be more effective than adaptive optimization applied to a poorly specified objective. 
Overall, these ablations show that performance gains arise from the interaction between offline structural knowledge and residual-based online updates. BO is an effective refinement mechanism only when this structure is present.

\subsection{Sensitivity Analysis}
\label{sec:results-sensitivity}

\begin{figure}[h]
\centering
\includegraphics[width=0.9\columnwidth]{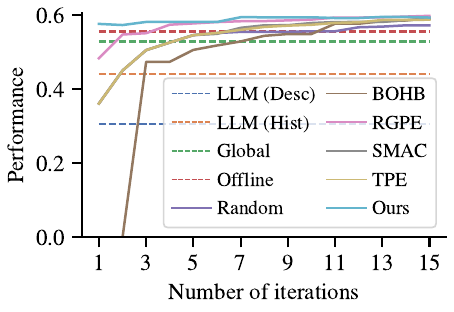}
\caption{Sensitivity of performance (average benchmark accuracy) to online evaluation budget.}
\label{fig:budget-curves}
\end{figure}

\noindent\textbf{Sensitivity to budget}. Figure~\ref{fig:budget-curves} plots performance against the number of online evaluations. For our method, online updates are driven by early-stopped SFT signals passed through the early-stop regressor, whereas the full-fidelity sequential baselines use full end-to-end evaluations as their optimization signal. Despite operating under this noisier low-cost feedback regime, our approach performs strongly even at low evaluation counts, due to rapid early gains from starting search in a favorable region identified by the offline ranker. With larger budgets, improvements become incremental, suggesting diminishing returns once the initial guidance has been locally refined. The mild non-monotonicity at very low budgets reflects noise in the proxy-feedback loop, but this effect is small and is quickly corrected as more evidence accumulates.

\vspace{0.05in}\noindent\textbf{Sensitivity to early-stop fidelity.} Table~\ref{tab:esfidelity} reports performance when early-stop runs are truncated at 25\%, 50\%, and 75\% of the full SFT trajectory. 
Very short truncation (25\%) degrades performance, suggesting that coarse early-stop signals are insufficiently reliable. Increasing truncation to 50\% yields substantial improvement, while extending to 75\% provides no consistent gains and can slightly hurt performance. Overall, signal utility appears to saturate beyond moderate fidelity, and performance remains stable across multiple truncation lengths. This robustness suggests that partial training dynamics are a practical proxy for online adaptation without requiring precise truncation tuning.

\begin{table}[h]
\centering
\small
\caption{Sensitivity to early-stop truncation length.}
\label{tab:esfidelity}
\begin{tabular}{cccc}
\toprule
\textbf{Variant} & \textbf{Performance} & \textbf{Regret} & \textbf{Cost} \\
\midrule
25\% & 0.561 $\pm$ 0.006 & 0.089 $\pm$ 0.006 & 8.7 $\pm$ 0.9 \\
50\% & 0.594 $\pm$ 0.002 & 0.056 $\pm$ 0.002 & 9.5 $\pm$ 0.4 \\
75\% & 0.585 $\pm$ 0.007 & 0.065 $\pm$ 0.007 & 10 $\pm$ 0.3 \\
\bottomrule
\end{tabular}
\end{table}

\vspace{0.05in}\noindent\textbf{Sensitivity to offline-guidance tempering.} We examined performance sensitivity to the annealing schedule of $w_t$, a coefficient that controls how strongly the online search remains anchored to the offline ranking. In Table~\ref{tab:wt}, we compare the default linear schedule $1.0 \to 0.7$ against constant $w_t=1.0$, constant $w_t=0.7$, and a more aggressive $1.0 \to 0.3$ schedule. Our default schedule achieves the strongest performance among the tested variants. Keeping offline contribution fixed at $1.0$ limits target-specific adaptation, whereas reducing its weight too strongly can discard useful transferable structure. Moderate annealing therefore balances preserving historical guidance early in search and allowing target-specific correction as online evidence accumulates.

\begin{table}[h]
\centering
\small
\caption{Sensitivity to offline-guidance tempering.}
\label{tab:wt}
\begin{tabular}{cccc}
\toprule
\textbf{$w_t$ Variant} & \textbf{Performance} & \textbf{Regret} & \textbf{Cost} \\
\midrule
$1.0 \to 0.7$ & 0.594 $\pm$ 0.002 & 0.056 $\pm$ 0.002 & 9.5 $\pm$ 0.4 \\
\midrule
$1.0$ & 0.545 $\pm$ 0.008 & 0.105 $\pm$ 0.008 & 9.6 $\pm$ 0.4 \\ 
$0.7$ & 0.530 $\pm$ 0.030 & 0.121 $\pm$ 0.030 & 9.5 $\pm$ 0.4 \\ 
$1.0 \to 0.3$ & 0.550 $\pm$ 0.009 & 0.100 $\pm$ 0.009 & 9.9 $\pm$ 0.1 \\ 
\bottomrule
\end{tabular}
\end{table}

\section{Conclusion}
AutoPipe’s core takeaway is that ranking transfer with local correction can effectively tune SFT–RL pipelines when full-fidelity trials are too expensive; an offline, dataset-conditioned ordering narrows the search to promising regions, and a small amount of online evidence helps correct dataset-specific miscalibration. In biomedical QA, in both our main LODO evaluation setting and a harder source-disjoint evaluation setting, this strategy consistently outperforms strong offline-only and early-stopped baselines, and remains competitive with full-fidelity HPO baselines while using substantially less online compute.

\section*{Limitations}
A remaining weakness is reliance on early-stopped proxy signals and historical priors: when truncated SFT dynamics are noisy or weakly related to post-RL outcomes, or under dataset shift, performance can be mildly non-monotonic at very low budgets before correcting as more evidence accrues. Additionally, our early-stop experiments use moderate SFT scales, and truncating at 50\% may remain costly for substantially larger training sets even if the RL stage dominates the overall cost. Evaluating more aggressive or absolute stopping rules in such settings is a promising direction for future work.

\section*{Acknowledgments}
This work was supported by the NAII Seed Fund A-8004350.

\bibliography{main}

\newpage
\appendix

\crefalias{section}{appendix}
\crefalias{subsection}{appendix}

\renewcommand{\thesection}{\Alph{section}}
\renewcommand{\thesubsection}{\thesection.\arabic{subsection}}
\counterwithin{figure}{section}
\counterwithin{table}{section}

\section{Problem Setup}
\label{appendix:problem}

\subsection{Dataset and Benchmark Details}
\label{appendix:dataset}

\subsubsection{Dataset details}
\label{appendix:dataset-details}

We study our methods on ten instruction-style datasets for medical reasoning post-training. The datasets differ along several axes, including prompt sources, teacher model, use of retrieval augmentation, and response format. Taken together, they form a heterogeneous testbed for investigating configuration selection in multi-stage LLM post-training pipelines for medical reasoning.

\medskip
\noindent \textbf{Datasets introduced in this paper}

\medskip
\noindent For the datasets newly introduced in this work, both reasoning traces and final responses are distilled from DeepSeek-R1, and examples are retained only if the final answer is correct. The datasets differ primarily in how prompts are constructed, as detailed below.

\paragraph{MQA.} MQA is built from a subset of medical and clinical questions sampled from the training splits of MedQA \citep{jin2021disease} and MedMCQA \citep{pal2022medmcqa}.

\paragraph{MQAR.} MQAR extends MQA with retrieval augmentation and is intended to assess whether a model can reason effectively in the presence of external evidence. Following the general spirit of MIRAGE \citep{chen2025benchmarking}, we assemble a specialized medical corpus from chunked medical textbooks provided by \citet{jin2021disease} together with PubMed abstracts collected by \citet{jin2023MedCPT}. For each question, we retrieve the top 10 passages from this corpus using BM25, and then distill both the response and reasoning trace conditioned on the retrieved evidence. The resulting examples therefore represent evidence-grounded clinical question answering.

\paragraph{PQA.} PQA is derived from the \texttt{pqa\_artificial} split of PubMedQA \citep{jin2019pubmedqa}. To broaden the range of contextual and linguistic settings seen during post-training, we construct three variants using distinct prompt templates. The first preserves the original question-context pairing, where the question is adapted from the article title and the context is the corresponding abstract. The second and third variants introduce a variable number of distractor abstracts retrieved by BM25 using the question as the query. In one variant, article titles are included before each abstract, emphasizing structured information extraction; in the other, titles are omitted, forcing greater semantic integration across the provided contexts. The final PQA dataset is obtained by mixing these three variants in equal proportions.

\paragraph{MQA(R).} MQA(R) is a composite dataset created by balanced sampling from MQA and MQAR.

\paragraph{MQA(R)PQA.} MQA(R)PQA is a composite dataset created by balanced sampling from MQA, MQAR, and PQA.

\medskip
\noindent \textbf{Open-source datasets from recent medical reasoning papers}

\paragraph{HuaTuo \citep{chen2024huatuogpt}.} HuaTuo is an open-ended medical instruction dataset constructed from difficult MedQA and MedMCQA examples reformatted into free-response form. The responses are distilled from GPT-4o. For SFT, we use the \texttt{en\_mix} subset of the \texttt{medical-o1-reasoning-SFT} split; for RL, we use the \texttt{medical-o1-verifiable-problem} split.

\paragraph{MedReason \citep{wu2025medreason}.} MedReason is designed to encourage explainable medical problem solving in LLMs through the use of medical knowledge graphs. For each example, relevant portions of a medical knowledge graph are retrieved and used to guide GPT-4o in synthesizing a reasoning chain. The underlying questions are drawn from a diverse mixture of medical and general benchmarks, including MedQA, MedMCQA, PubMedQA, MMLU, MedXpertQA, and Humanity’s Last Exam.

\paragraph{MedS3 \citep{jiang2025meds}.} MedS3 employs a self-evolution framework based on Monte Carlo Tree Search to generate and verify long-form reasoning trajectories. We use the resulting MedS3 dataset, whose prompts come from 16 medical and general-domain datasets and are paired with trajectories produced by the MedS3 pipeline.

\paragraph{ReasonMed \citep{sun2025reasonmed}.} ReasonMed is produced by a multi-agent pipeline that generates, verifies, and iteratively revises reasoning trajectories distilled from multiple teacher models. Its curation pipeline reduces an initial pool of 1.75 million trajectories to 370k high-quality examples. The prompts are sourced from MedQA, MedMCQA, MMLU, and PubMedQA.

\paragraph{m23k \citep{huang2026m1}.} m23k is created through a multi-stage refinement process that includes difficulty filtering (to remove instances already solvable by standard base models), reasoning generation with DeepSeek-R1, and filtering based on final-answer correctness. Its prompts come from MedMCQA, MedQA, HeadQA, and PubMedQA.

\paragraph{CasiMedicos-Arg \citep{sviridova-etal-2024-casimedicos}.} CasiMedicos-Arg is a medical QA dataset derived from Spanish MIR medical residency examinations (which are not represented among the previous ten datasets). We use the English configuration released on HuggingFace, which contains 622 examples across data splits. Because CasiMedicos-Arg is used here as a post-training target rather than as a downstream evaluation benchmark, we combine all data splits before preprocessing. Owing to the smaller size of CasiMedicos-Arg, we partition the available examples into a similar 5:2 ratio for SFT:RL allocation, rather than matching the fixed sizes used for the primary LODO targets.
\medskip

\noindent We apply the same preprocessing pipeline to all datasets, consisting of length filtering followed by sampling. For datasets introduced in this paper, we explicitly enforce balanced representation across prompt sources in the final dataset (for example, MQA contains 50\% MedQA prompts and 50\% MedMCQA prompts). For open-source datasets from prior work, we instead perform stratified sampling according to the original source distribution, unless a single source overwhelmingly dominates the pool (e.g., more than 90\% of examples). In such cases, we switch to equal sampling across all available sources in order to preserve source diversity.

\subsubsection{Benchmark details}
\label{appendix:benchmark-details}

Final model performance is measured on twelve biomedical question-answering benchmarks. These benchmarks span different answer formats, domain scopes, and difficulty levels, allowing us to evaluate generalization across both biomedical knowledge recall and clinical reasoning.

\paragraph{BioASQ-Y/N \citep{tsatsaronis2015overview, krithara2023bioasq}.} Following \citet{xiong2024benchmarking}, we use a modified version of the BioASQ Task B machine reading comprehension benchmark. The benchmark contains 618 binary (Yes/No) questions drawn from the 2019-2023 competition cycles. In contrast to the original task formulation, which provides gold literature snippets, the modified version removes these supporting snippets, requiring the model to answer using only parametric biomedical knowledge.

\paragraph{GPQA (Biology) \citep{rein2024gpqa}.} This benchmark consists of graduate-level biology multiple-choice questions authored by domain experts.

\paragraph{MedBullets \citep{chen2025benchmarking}.} MedBullets is a recently introduced medical licensing exam-style benchmark composed of multiple-choice questions based on realistic clinical scenarios. \citet{chen2025benchmarking} report that it is more difficult than older benchmarks such as MedQA, likely due in part to its more recent curation. We evaluate both the 4-option and 5-option versions, with the latter adding an extra distractor choice to increase difficulty.

\paragraph{MedMCQA \citep{pal2022medmcqa}.} MedMCQA is a large-scale multiple-choice medical QA benchmark derived from Indian medical entrance examinations.

\paragraph{MedQA \citep{jin2021disease}.} MedQA contains medical licensing examination questions collected from multiple geographic regions. We use the US subset.

\paragraph{MedXpertQA \citep{zuo2025medxpertqa}.} MedXpertQA is curated from medical professional and specialty board examinations through a pipeline involving AI-based filtering, question rewriting, and physician review. Each question has 10 answer options, which is substantially more than is typical in medical QA benchmarks. In addition, the questions cover a broad range of body systems and specialties, making this benchmark relatively challenging.

\paragraph{MMLU (Medicine) \citep{hendrycks2021measuring}.} We use the six medicine-related MMLU subjects: anatomy, clinical knowledge, college biology, college medicine, medical genetics, and professional medicine.

\paragraph{MMLU-Pro (Biology \& Health) \citep{wang2024mmlu}.} MMLU-Pro is a more challenging variant of MMLU. We use its biology and health subsets.

\paragraph{PubMedQA \citep{jin2019pubmedqa}.} PubMedQA is a biomedical QA benchmark in which PubMed abstracts are paired with question statements derived from article titles, and the task is to determine whether the abstract entails the statement (yes/no/maybe).

\paragraph{PubMedQA (Questions Only).} In the \textbf{questions-only} setting, the model receives the question statement without any accompanying abstract. This isolates parametric biomedical knowledge.

\paragraph{PubMedQA (With Distractor Abstracts).} In the \textbf{distractor} setting introduced in this work, each question is paired with its gold abstract along with several distractor abstracts. This setting requires the model to identify useful evidence in addition to producing the correct answer. Distractor abstracts are retrieved with BM25 using the question statement as the query.

\medskip
\noindent To keep evaluation costs tractable, we evaluate on fixed subsets rather than the complete benchmark suites. Unless stated otherwise, we randomly sample 100 examples per benchmark from the test split, or from the validation split if no test split is available. Two benchmarks are exceptions to this default: GPQA (Biology), which contains 78 examples, and MMLU (Medicine), which contains 123 examples after aggregating the six medicine-related subjects. Altogether, the final evaluation suite contains 1,201 prompts.

\subsection{Sampling from the Configuration Space}
\label{appendix:configspace}

To build the offline corpus, we sample 600 configurations from the search space using a scheme that guarantees equal coverage of every possible value of each hyperparameter. This design avoids concentrating too heavily on a narrow region of the search space and increases the diversity of the training data available to both the offline ranker and the early-stop regressor.

We further increase diversity in the offline corpus by varying dataset scale across training stages. For SFT, we use subsets of 1000, 2000, and 4000 samples. For RL, we vary the training horizon across 50, 100, and 200 steps, adjusting the amount of RL data accordingly. As a result, the offline corpus captures variation not only in hyperparameter settings but also in the scale of the underlying training procedure, thereby exposing downstream models to a broader range of optimization behaviors.

\section{Methodological Details}
\label{appendix:method}

\subsection{Offline Surrogate Model}
\label{appendix:offline}

This section provides additional implementation details for the offline ranker that provides guidance during online optimization.

\subsubsection{Feature Construction and Selection}

For each dataset-configuration pair, we construct a heterogeneous feature representation that jointly describes the configuration and the dataset. The full initial feature set is divided into four groups:

\begin{itemize}[nolistsep]
    \item \textbf{Configuration features (13).} These represent the post-training hyperparameter choices, including learning rates, batch sizes, numbers of epochs, and other discrete design decisions. They correspond to the configuration space in Table~\ref{tab:configspace}, with the base model one-hot encoded.
    \item \textbf{Simple dataset features (26).} These are coarse descriptive statistics such as token counts, vocabulary size, mean sequence length, and average embedding distances, intended to summarize basic structural properties of each dataset.
    \item \textbf{Rich dataset features (82).} These capture higher-level linguistic and structural properties, such as lexical diversity, type--token ratio, and measures of instructional or textual complexity.
    \item \textbf{Dataset overlap features (204).} These quantify similarity between the training dataset and a set of target benchmarks using several similarity measures, including TF-IDF cosine similarity, embedding-based similarity, and Jaccard overlap.
\end{itemize}

\medskip
\noindent All of these features are available before any post-training run is executed and do not depend on labels or post hoc evaluation signals from target datasets. To reduce redundancy and improve robustness, we apply a dataset-aware feature filtering procedure before training. For each dataset, we flag features whose variance falls below $10^{-2}$ times the median feature variance, as well as features that are highly correlated (Pearson correlation $\geq 0.97$). Any feature flagged in at least 9 of the 10 datasets is removed. After filtering, the dimensionality decreases from 325 to 223 features, consisting of 13 configuration features, 22 simple dataset features, 59 rich dataset features, and 129 dataset overlap features.

\subsubsection{Training and Validation Protocol}

We train and evaluate the ranker under a leave-one-dataset-out (LODO) protocol in order to match the intended deployment setting. Each dataset is treated as a ranking group, with all associated configurations forming a ranked list. We also treat multiple subsets sampled from the same underlying dataset as belonging to the same ranking group, since their computed dataset characteristics are highly similar. In each fold, the model is trained on nine datasets and evaluated on the held-out tenth dataset.

We use XGBRanker as the base model because it can represent non-linear interactions between dataset features and configuration features while directly optimizing a ranking objective. To improve stability, we train an ensemble of three XGBRanker models with identical hyperparameters but different random seeds. Final scores and evaluation metrics are computed from the average of the ensemble predictions.

Hyperparameters are selected by grid search over standard XGBoost settings such as tree depth, learning rate, and number of estimators. We choose the final setting based on both the mean and minimum NDCG@5 across datasets so as to favor robust cross-dataset performance rather than strong average performance alone.

\subsection{Early-Stop Regressor for Low-Cost Online Updates}
\label{appendix:early_stop_regressor}

This section provides implementation details for the early-stop regressor used in the online phase. In contrast to the offline ranker, whose scores can be computed without running new experiments, the early-stop regressor operates on partial signals observed from a small number of executed runs. Its purpose is to produce a low-cost proxy signal that can be used to update the online optimizer without repeatedly paying the cost of full RL training.

\subsubsection{Feature Construction from SFT Training Dynamics}

The regressor uses training dynamics recorded during SFT. We focus on SFT because it is much cheaper than RL while still yielding informative signals about downstream behavior: the eventual RL outcome depends heavily on the quality and stability of the SFT initialization, so early SFT dynamics can provide useful evidence about which configurations are likely to perform well in the full post-training pipeline.

For each run, we extract logged time-series metrics such as training loss, validation loss, gradient norm statistics, entropy-like quantities, and related training signals. From each trajectory, we compute summary statistics intended to capture both overall behavior and local stability. These include the mean, maximum, minimum, standard deviation, slope, and mean difference.

To avoid leaking information from beyond the early-stop point, all training-dynamic features are computed only on truncated prefixes of the trajectory. We define a \emph{prefix window} covering the first 50\% of the SFT trajectory. We also define a \emph{late window} corresponding to the last 20\% of that prefix window, which is intended to capture late-stage stability and local convergence behavior. For every metric and summary statistic, we compute features over both windows.

\subsubsection{Feature Filtering}

As with the offline ranker, we prune features to reduce redundancy and improve generalization. We remove features with near-zero variance (below $10^{-2}$ times the median feature variance) and features with high pairwise correlation (Pearson correlation $\geq 0.97$). The same pruning rule is applied across datasets to reduce the risk of overfitting to logging artifacts specific to any one dataset.

\subsubsection{Regression Target: Standardized Residual Correction}

An important design choice is the regression target. The offline ranker yields a dataset-conditioned ordering over configurations, but its score is only an imperfect proxy for true end-to-end performance. At the same time, directly regressing on absolute performance is undesirable because the target may be dominated by dataset-level shifts in performance scale, i.e., some datasets may simply be easier or harder than others. Because the regressor is used to update the online procedure from noisy partial observations, we instead define the target as a \emph{standardized residual correction}. Let $y_{d,c}$ denote the full end-to-end performance of configuration $c$ on dataset $d$, and let $s_{d,c}^{(z)}$ denote the within-dataset z-normalized score assigned by the offline ranker. We first z-normalize the end-to-end performance:
\[
y_{d,c}^{(z)} = \frac{y_{d,c} - \mu_d}{\sigma_d + \epsilon}
\]
Here, $\mu_d$ and $\sigma_d$ denote the normalization statistics used for dataset $d$ in that fold; for training datasets they are estimated from the offline corpus, while the held-out dataset is never used to fit these statistics. We then define the regression target as
\[
r_{d,c} = y_{d,c}^{(z)} - s_{d,c}^{(z)}
\]
The regressor is trained to predict $r_{d,c}$ from configuration features, dataset features, and early-stop dynamic features only. In particular, the offline score $s_{d,c}^{(z)}$ is not supplied directly as an input feature.

\paragraph{Inference and evaluation.} At inference time, we reconstruct the final score used for ranking or configuration selection by anchoring to the offline ranker and adding the predicted correction $\hat{r}_{d,c}$:
\[
\hat{y}^{(z)}_{d,c} = s_{d,c}^{(z)} + \hat{r}_{d,c}
\]
This makes the regressor a \emph{conservative correction} mechanism: if the early-stop signals are weak or noisy, predictions remain close to the offline score, reducing the chance that BO is misled by unreliable pseudo-observations. For evaluation, we compare $\hat{y}^{(z)}_{d,c}$ against the true standardized performance $y^{(z)}_{d,c}$ using ranking and selection metrics.

\subsubsection{Training and Validation Protocol}

We use XGBRegressor models to map early-stop features to residual corrections. The regressor is trained and evaluated under the same LODO protocol as the offline ranker. To improve stability and obtain a crude uncertainty estimate for online BO updates, we train an ensemble of five regressors with identical hyperparameters but different random seeds. Their predictions are averaged, and the ensemble spread is used as a noise estimate when the regressor outputs are incorporated as pseudo-observations in the online phase. Hyperparameters are selected by grid search over standard XGBoost settings. Final model selection is based on the mean and minimum NDCG@5 across datasets.

\subsection{Online Implementation}
\label{appendix:online}

This section describes how post-training configurations are selected during the online phase. The BO loop combines prior guidance from the offline ranker with sequential updates derived from partial training signals and the early-stop regressor.

\subsubsection{Gaussian Process over Residual Corrections}

For each dataset--configuration pair $(d,c)$, the offline ranker provides a standardized score $s_{d,c}^{(z)}$. As described above, this score induces a dataset-conditioned ordering over configurations and is available before any online evaluation is performed.

Rather than modeling absolute performance directly, we again work in standardized residual space, analogous to the early-stop regressor. In the online setting, however, we additionally introduce a dataset-specific offset term to improve adaptation. Concretely, we define
\[
\tilde{r}_{d,c} = y_{d,c}^{(z)} - s_{d,c}^{(z)} - b_d
\]
where $y_{d,c}^{(z)}$ is standardized end-to-end performance, $s_{d,c}^{(z)}$ is the standardized offline ranker score, and $b_d$ is a dataset-specific offset. The GP then models a latent function $f_d(c)$ such that
\[
f_d(c) \sim \mathcal{GP}(0, k(c,c'))
\]
By construction, the GP therefore has zero mean in residual space. We use a kernel with automatic relevance determination (ARD) across feature dimensions, allowing different features and hyperparameters to operate on different characteristic length scales. This is desirable because different components of the post-training pipeline can influence optimization behavior at very different granularities.

\subsubsection{Updating the GP}
\label{appendix:updateGP}

\paragraph{Warm-start Initialization.}
We initialize BO with a diverse set of configurations chosen using the offline ranker scores. The goal is to improve early coverage of the search space and reduce the risk of premature convergence.

\paragraph{Acquisition Function.}
At BO iteration $t$, let $\mu_{f,t}(c)$ and $\sigma_{f,t}(c)$ denote the GP posterior mean and standard deviation for the residual correction. The standardized performance estimate is
\[
\hat{y}_{d,c,t}^{(z)} = w_t\, s_{d,c}^{(z)} + b_{d,t} + \mu_{f,t}(c)
\]
where $b_{d,t}$ is the current estimate of the dataset-specific offset and $w_t \in [0,1]$ is a tempering coefficient that controls how strongly the search is anchored to the offline ranker. We linearly anneal $w_t$ from $1.0$ to $0.7$, so that earlier iterations rely more heavily on transferable offline structure, while later iterations are allowed to adapt more strongly to dataset-specific evidence. We use an Upper Confidence Bound (UCB) acquisition function:
\[
\alpha_t(c) = \hat{y}_{d,c,t}^{(z)} + \beta_t\, \sigma_{f,t}(c)
\]
where $\beta_t$ governs the exploration--exploitation trade-off. Because the offline score does not vary across BO iterations, exploration is driven by uncertainty in the residual correction, while the overall ranking remains tied to the offline guidance. We use a decaying schedule for $\beta_t$, starting with stronger exploration and gradually shifting toward exploitation as evidence accumulates. 

\paragraph{Early-Stop Updates.}
At each BO iteration, the selected configuration is partially evaluated by running SFT to a predefined early-stop point. The resulting training dynamics are fed into the early-stop regressor (Appendix~\ref{appendix:early_stop_regressor}), which outputs a predicted pre-bias residual correction $\hat{r}_{d,c}$. We store these pseudo-observations in their pre-bias form. After updating the dataset-specific offset $b_{d,t}$, the GP is refit on centered targets:
\[
\hat{r}_{d,c} - b_{d,t}= f_d(c) + \eta, \quad \eta \sim \mathcal{N}(0, \sigma^2_{\text{ens}})
\]
where $\sigma^2_{\text{ens}}$ is the ensemble variance of the regressor prediction, subject to a minimum variance floor. This separation allows the offset term to capture dataset-wide miscalibration while the GP models configuration-specific deviations.

\paragraph{Dataset-specific Offset.} 
To account for dataset-level mismatch between offline guidance and target-dataset behavior, we maintain a dataset-specific offset $b_d$. Let $\Delta_i$ denote the $i$-th stored pre-bias residual pseudo-observation. After $t$ online observations have been collected, we update the offset using the inverse-variance-weighted mean
\[
b_{d,t} = \frac{\sum_{i=1}^{t} \Delta_i / v_i}{\sum_{i=1}^{t} 1 / v_i},
\]
where $v_i$ is the observation variance associated with $\Delta_i$. The GP at iteration $t$ is then refit on centered targets $(\Delta_i - b_{d,t})$, so that $b_{d,t}$ absorbs dataset-wide shift while the latent function $f_d(c)$ models configuration-specific departures.

\section{Evaluation Methodology}
\label{appendix:eval}

\subsection{End-to-end Evaluation}
\label{appendix:e2eeval}

To ensure a controlled comparison across LODO splits, we define a fixed evaluation-scale version of each of the ten held-out datasets. Specifically, we fix the number of SFT prompts across target datasets and likewise fix the number of RL prompts. We refer to these fixed-scale dataset versions as the \emph{canonical distributions}. Dataset meta-features are computed on these canonical distributions, and all methods and baselines are evaluated under the same data configuration to ensure a fair comparison.

\medskip
\noindent All training workflows are implemented using the HuggingFace library, with \texttt{SFTTrainer} for supervised fine-tuning and \texttt{GRPOTrainer} for reinforcement learning. For SFT, we use a cosine learning-rate schedule with warmup ratio $0.1$. For GRPO, we use the \texttt{dr\_grpo} loss with $\epsilon_{\text{high}} = 0.28$. Across all tasks, the reward is defined as a composite rule-based signal consisting of:
\begin{itemize}[nolistsep]
    \item Format reward: a small incentive for following the requested output structure
    \item Accuracy reward: a binary reward based on exact match with the ground-truth answer
\end{itemize}
To enable automatic answer parsing, we append the instruction ``At the end of your response, output \texttt{\{answer\_format\}} where X is your chosen answer.'' to all RL prompts. To make the model less sensitive to a single answer-format instruction, we vary \texttt{\{answer\_format\}} across prompts using eight distinct formats, including \texttt{\textbackslash boxed\{X\}} and \texttt{**Answer: X**}.

\medskip
\noindent We conduct all evaluations with the \textbf{lm-evaluation-harness} framework, which provides a standardized interface across heterogeneous tasks. All benchmarks use the same prompt template (Prompt~\hyperref[prompt:eval]{1}) and require a free-form response ending in a final answer. In post-processing, we apply a regex-based filter to extract the first short-form answer token following the delimiter ``Answer:''. We then compute case- and punctuation-insensitive exact-match accuracy against the gold label. Generation settings are kept consistent across benchmarks, including stochastic decoding, a fixed maximum generation length, and standard end-of-sequence stopping rules. Because decoding is stochastic, all benchmark results are averaged over three random seeds.

\begin{tcolorbox}[
  title=Prompt 1: Evaluation prompt template,
  colback=gray!5,
  colframe=black,
  boxrule=0.5pt,
  arc=2pt,
  left=6pt,
  right=6pt,
  top=6pt,
  bottom=6pt
]
\label{prompt:eval}
Solve the following question step-by-step. At the end of your answer, output "Answer: X" where X is the option corresponding to your final answer.

\medskip
\{question\} \\
A: ... \\
B: ... \\
C: ... \\
D: ... \\
(additional options if any)
\end{tcolorbox}

\paragraph{Repeated runs and sources of randomness.} In addition to decoding stochasticity, several configuration selection methods contain stochastic components. We therefore repeat each stochastic method thrice using independent experimental seeds. For methods with learned offline components, the corresponding gradient-boosted models are retrained under each repetition with independently seeded model initialization (this is distinct from the multiple independently seeded models within each offline ranker/regressor ensemble). Repetition of sequential methods rerun the complete online selection procedure with an independent search seed. The final configuration selected in each repetition is also evaluated using the three-seed protocol described above. Hence, each repetition captures the stochasticity of the method components relevant to the corresponding approach, rather than only that of the final benchmark evaluation. We report performance, regret, and cost as the mean and standard error across the three independent repetitions.

\subsection{Baseline Implementations}
\label{appendix:baseline}

This section describes the implementation details of the baselines.

\subsubsection{Static \& Zero-Shot Baselines}

\paragraph{Random (Single run).} We uniformly sample one configuration from the search space and execute it.

\begin{figure*}[p]
\begin{mdframed}[backgroundcolor=gray!5, roundcorner=2pt, linecolor=black, linewidth=0.5pt, innerleftmargin=15pt, innerrightmargin=15pt, innertopmargin=10pt, innerbottommargin=10pt]
\small
\textbf{Prompt 2: LLM Recommendation (Description-only) Template}
\vspace{0.4em}
\hrule
\vspace{0.8em}

\begin{flushleft}
You are an expert researcher executing an LLM post-training workflow consisting of supervised fine-tuning (SFT) followed by reinforcement learning (RL). You are tasked with finding a suitable hyperparameter configuration for this workflow. You are provided with the objective of the post-training workflow, the configuration space, and some characteristics about the dataset you will be using.

\textbf{\#\# Objective}

The outcome of the post-training should be a medical reasoning model. Its performance will be evaluated on a series of medical benchmarks, including MedQA, MedMCQA, PubMedQA, MedBullets, MMLU (Med), MMLU-Pro (Biology/Health), GPQA (Biology), MedXpertQA and BioASQ. You should aim to maximize performance on this set of benchmarks. You cannot run experiments or use historical results; decide only from the information given.

\textbf{\#\# Configuration space}
\begin{itemize}[noitemsep,topsep=2pt,leftmargin=1.5em]
    \item \texttt{base\_model}: \{L1 (Llama-3.2-1B-Instruct), L3 (Llama-3.2-3B-Instruct), L8 (Llama-3.1-8B-Instruct), Q1 (Qwen2.5-1.5B-Instruct), Q3 (Qwen2.5-3B-Instruct), Q8 (Qwen2.5-7B-Instruct)\}
    \item \texttt{sft\_lr}: \{$1\times10^{-5}, 2\times10^{-5}, 5\times10^{-5}$\}, \texttt{sft\_batch}: \{32, 64, 128\}, \texttt{sft\_epoch}: \{1, 2, 3\}
    \item \texttt{rl\_lr}: \{$1\times10^{-6}, 2\times10^{-6}, 5\times10^{-6}, 1\times10^{-5}$\}, \texttt{rl\_batch}: \{64, 128, 256\}, \texttt{rl\_beta}: \{0, 0.05, 0.1\}, \texttt{rl\_rollouts}: \{8, 16\}, \texttt{rl\_temp}: \{0.7, 0.9, 1.1\}
\end{itemize}

\textbf{\#\# Dataset characteristics} \\
\textit{SFT Stage:}
\begin{itemize}[noitemsep,topsep=0pt,leftmargin=1.5em]
    \item prompt sources: $\langle \text{src\_sft} \rangle$
    \item n\_samples: $\langle \text{data\_n\_sft\_samples} \rangle$
    \item prompt\_len\_tokens: mean=$\langle \mu_{p,sft} \rangle$, sd=$\langle \sigma_{p,sft} \rangle$
    \item completion\_len\_tokens: mean=$\langle \mu_{c,sft} \rangle$, sd=$\langle \sigma_{c,sft} \rangle$
    \item completion\_source: $\langle \text{comp\_src} \rangle$
    \item task\_format: $\langle \text{fmt\_sft} \rangle$
    \item num\_choices\_min: $\langle \text{min\_c} \rangle$, max: $\langle \text{max\_c} \rangle$
\end{itemize}
\textit{RL Stage:}
\begin{itemize}[noitemsep,topsep=0pt,leftmargin=1.5em]
    \item prompt sources: $\langle \text{src\_rl} \rangle$
    \item n\_samples: $\langle \text{data\_n\_rl\_samples} \rangle$
    \item prompt\_len\_tokens: mean=$\langle \mu_{p,rl} \rangle$, sd=$\langle \sigma_{p,rl} \rangle$
    \item task\_format: $\langle \text{fmt\_rl} \rangle$
    \item num\_choices\_min: $\langle \text{min\_c} \rangle$, max: $\langle \text{max\_c} \rangle$
    \item reward\_type: accuracy (exact match)
\end{itemize}

\textbf{\#\# Your recommendation}

Based on the above configuration space and task information, recommend a suitable hyperparameter configuration. All hyperparameter values must be chosen from the configuration space. Format your recommendation in a single JSON object:
\end{flushleft}

\begin{verbatim}
{
    "base_model": "...",
    "sft_lr": "...",
    "sft_batch": "...",
    "sft_epoch": "...",
    "rl_lr": "...",
    "rl_batch": "...",
    "rl_beta": "...",
    "rl_rollouts": "...",
    "rl_temp": "..."
}
\end{verbatim}
\end{mdframed}
\caption{Prompt template used for LLM Recommendation (Description-only) baseline. Values enclosed in angled brackets are dependent on the unseen dataset.}
\label{fig:llm_desc_template}
\end{figure*}

\begin{figure*}[p]
\begin{mdframed}[backgroundcolor=gray!5, roundcorner=2pt, linecolor=black, linewidth=0.5pt, innerleftmargin=15pt, innerrightmargin=15pt, innertopmargin=10pt, innerbottommargin=10pt]
\small
\textbf{Prompt 3: LLM Recommendation (History-conditioned) Template}
\vspace{0.4em}
\hrule
\vspace{0.8em}

You are an expert researcher executing an LLM post-training workflow consisting of supervised fine-tuning (SFT) followed by reinforcement learning (RL). You are tasked with finding a suitable hyperparameter configuration for this workflow. You are provided with the objective of the post-training workflow, the configuration space, some characteristics about the dataset you will be using, and the results of some previous experiments using similar datasets.

\textbf{\#\# Objective} \\
The outcome of the post-training should be a medical reasoning model. Its performance will be evaluated on a series of medical benchmarks, including MedQA, MedMCQA, PubMedQA, MedBullets, MMLU (Med), MMLU-Pro (Biology/Health), GPQA (Biology), MedXpertQA and BioASQ. You should aim to maximize performance on this set of benchmarks. You cannot run new experiments; decide only from the information given.

\textbf{\#\# Configuration space} 
\begin{itemize}[noitemsep,topsep=2pt,leftmargin=1.5em]
    \item \texttt{base\_model}: \{L1 (Llama-3.2-1B-Inst), L3 (Llama-3.2-3B-Inst), L8 (Llama-3.1-8B-Inst), Q1 (Qwen2.5-1.5B-Inst), Q3 (Qwen2.5-3B-Inst), Q8 (Qwen2.5-7B-Inst)\}
    \item \texttt{sft\_lr}: \{$1\times10^{-5}, 2\times10^{-5}, 5\times10^{-5}$\}, \texttt{sft\_batch}: \{32, 64, 128\}, \texttt{sft\_epoch}: \{1, 2, 3\}
    \item \texttt{rl\_lr}: \{$1\times10^{-6}, 2\times10^{-6}, 5\times10^{-6}, 1\times10^{-5}$\}, \texttt{rl\_batch}: \{64, 128, 256\}, \texttt{rl\_beta}: \{0, 0.05, 0.1\}, \texttt{rl\_rollouts}: \{8, 16\}, \texttt{rl\_temp}: \{0.7, 0.9, 1.1\}
\end{itemize}

\textbf{\#\# Dataset characteristics} \\
\textit{SFT Stage:}
\begin{itemize}[noitemsep,topsep=0pt,leftmargin=1.2em]
    \item prompt sources: $\langle \text{sft\_prompt\_sources} \rangle$
    \item n\_samples: 2000
    \item prompt\_len\_tokens: mean=$\langle \mu_{p,s} \rangle$, sd=$\langle \sigma_{p,s} \rangle$
    \item completion\_len\_tokens: mean=$\langle \mu_{c,s} \rangle$, sd=$\langle \sigma_{c,s} \rangle$
    \item completion\_source: $\langle \text{teacher\_model} \rangle$
    \item task\_format: $\langle \text{task\_format} \rangle$
\end{itemize}
\textit{RL Stage:}
\begin{itemize}[noitemsep,topsep=0pt,leftmargin=1.2em]
    \item prompt sources: $\langle \text{rl\_prompt\_sources} \rangle$
    \item n\_samples: 800
    \item prompt\_len\_tokens: mean=$\langle \mu_{p,r} \rangle$, sd=$\langle \sigma_{p,r} \rangle$
    \item task\_format: $\langle \text{task\_format} \rangle$
    \item reward\_type: accuracy (exact match)
\end{itemize}

\textbf{\#\# Nearest neighbor instances (historical results from other datasets)} 

\begin{mdframed}[ backgroundcolor=white, roundcorner=2pt, linecolor=gray!60, linewidth=0.5pt, innerleftmargin=8pt, innerrightmargin=8pt, innertopmargin=6pt, innerbottommargin=6pt ]
\textbf{Neighbor instance $i$ (similarity=rank $i$/9 among training datasets)}
\begin{itemize}[noitemsep,topsep=2pt,leftmargin=1.2em]
    \item \textbf{Dataset summary:} 
    \begin{itemize}[noitemsep]
        \item Prompt sources: $\langle \text{src}_{i} \rangle$, Task format: $\langle \text{fmt}_{i} \rangle$
        \item SFT: $n=\langle n_{s,i} \rangle$, prompt\_len\_mean=$\langle \mu_{p,s,i} \rangle$, completion\_source=$\langle \text{teacher}_{i} \rangle$
        \item RL: $n=\langle n_{r,i} \rangle$, prompt\_len\_mean=$\langle \mu_{p,r,i} \rangle$
    \end{itemize}
    \item \textbf{Config 1:} \texttt{\{"base\_model": "...", "sft\_lr": ..., "rl\_lr": ..., ...\}} [\text{acc}=$\langle \text{acc}_{i,1} \rangle$]
    \item \textbf{Config 2:} \texttt{\{"base\_model": "...", "sft\_lr": ..., "rl\_lr": ..., ...\}} [\text{acc}=$\langle \text{acc}_{i,2} \rangle$]
\end{itemize}
\end{mdframed}
\begin{center}
$\dots$ \textit{(Repeated for top-5 neighbors)} $\dots$ 
\end{center}

\textbf{\#\# Your recommendation} \\
Based on the above configuration space and task information, recommend a suitable hyperparameter configuration. Format your recommendation in a single JSON object:

\begin{verbatim}
{
    "base_model": "...",
    "sft_lr": "...",
    "sft_batch": "...",
    "sft_epoch": "...",
    "rl_lr": "...",
    "rl_batch": "...",
    "rl_beta": "...",
    "rl_rollouts": "...",
    "rl_temp": "..."
}
\end{verbatim}
\end{mdframed}
\caption{The full history-conditioned prompt template, utilizing meta-feature-based retrieval to provide historical experiment results (nearest neighbors) to the LLM surrogate.}
\label{fig:llm_hist_template}
\end{figure*}

\paragraph{LLM Recommendation (Description-only).} For this baseline, we query GPT 5.2 using the prompt shown in Figure~\ref{fig:llm_desc_template} and obtain a single recommended configuration for each dataset. The prompt includes only the task description together with a basic summary of the unseen dataset.

\paragraph{LLM Recommendation (History-conditioned).} For this baseline, we query GPT 5.2 using the prompt in Figure~\ref{fig:llm_hist_template} and obtain one recommendation per dataset. To provide historical context, we retrieve neighboring datasets by cosine similarity to the target dataset in the joint meta-feature space.

\paragraph{Global.} For each held-out dataset, we define the ``global best'' configuration as the single hyperparameter setting with the highest absolute performance across the other nine datasets. That configuration is then evaluated on the held-out dataset. This baseline therefore benefits from prior experimental evidence, but it does not adapt to target-dataset-specific properties such as training size, which may vary across the offline corpus.

\paragraph{Offline-only (Ranker).} This baseline measures the zero-shot transfer ability of the offline ranker. For each held-out dataset, the ranker scores all $34{,}992$ configurations in the search space. We then run a single full-fidelity evaluation on the highest-ranked configuration and report its final accuracy.

\subsubsection{Sequential Baselines}

With the exception of BOHB, all sequential baselines are evaluated in both full-training and early-stop variants. The full-training variants execute the entire pipeline (SFT, RL, and evaluation) and use the resulting end-to-end accuracy as the optimization signal. The early-stop variants instead rely on a separate regressor trained to predict final performance $y_{d,c}$ directly, rather than residual corrections, since these baselines do not have access to the offline ranker used by our method. Every sequential method is given a fixed budget of 15 evaluations. SMAC and TPE are warm-started with the first five configurations sampled by Random Search. In the full-training setting these initial points use ground-truth scores, whereas in the early-stop setting they use regressor-predicted scores.

\paragraph{Random Search.} We sample 15 configurations uniformly at random from the search space and evaluate them.

\paragraph{BOHB \citep{falkner2018bohb}.} We use the official \texttt{HpBandSter} implementation. The budget $b$ is defined in terms of RL-stage training fidelity because SFT contributes relatively little to total computational cost. We set the minimum and maximum budgets to $b_{\min}=1$ and $b_{\max}=4$, corresponding respectively to 25\% and 100\% of the full RL training horizon, and use elimination rate $\eta=2$. Note that BOHB’s fidelity notion is defined over RL-stage training budget, whereas our method uses SFT-truncated runs combined with an early-stop predictor. The two are therefore not identical notions of low-fidelity evaluation, but each reflects the most natural low-cost signal available to the corresponding method.

\paragraph{RGPE \citep{feurer2018scalable}.} We implement RGPE following the BoTorch reference implementation. 

\paragraph{SMAC \citep{hutter2011sequential}.} We use the SMAC3 framework \citep{Lindauer_SMAC3_A_Versatile_2022}, version 2.3.1.

\paragraph{TPE \citep{bergstra2011algorithms}.} We use the Optuna implementation \citep{optuna_2019}, version 4.7.0, with \texttt{TPESampler} in multivariate mode so as to better capture dependencies among hyperparameters.

\section{Extended Results}
\label{appendix:extresults}

This section provides the dataset-based breakdowns for all aggregated results presented in the main paper. 

\subsection{End-to-End Configuration Selection Performance}
\label{appendix:results-e2e}

Tables~\ref{tab:perf-expanded}, ~\ref{tab:regret-expanded} and~\ref{tab:cost-expanded} report dataset-level performance, regret and cost. Although certain datasets prove harder than others, the results overall demonstrate that AutoPipe remains competitive across the held-out datasets, while requiring substantially lower cost than full-evaluation sequential baselines. 

In particular, an informative pattern emerges when comparing AutoPipe directly with the \emph{Offline-only} baseline. AutoPipe improves upon the offline recommendation on majority of datasets, but the magnitude of improvement varies substantially; gains are negligible on datasets for which the transferred ranking is already strong, whereas considerably larger improvements occur on folds where the offline recommendation is weaker. This behavior is consistent with the intended role of the online phase: rather than uniformly overriding the guidance of the offline phase, it provides greatest benefit when dataset-specific correction is most necessary. 

\begin{table*}[p]
\footnotesize
\setlength{\tabcolsep}{2pt}
\renewcommand{\arraystretch}{0.95}
\caption{
Configuration selection performance across ten held-out datasets under a fixed
number of online evaluations. Performance is calculated as the average benchmark
accuracy. Results for repeated methods are reported as mean $\pm$ standard error
over three independent repetitions. Bold indicates the best performance per dataset;
italics indicate the second-best.
}
\label{tab:perf-expanded}

\begin{tabularx}{\textwidth}{@{}MYYYYYY@{}}

\cmidrule[\heavyrulewidth](r){1-6}

& \multicolumn{5}{c}{\textbf{Datasets 1--5}} & \\
\cmidrule(lr){2-6}

\textbf{Method}
& \multicolumn{1}{c}{\footnotesize\textbf{1}}
& \multicolumn{1}{c}{\footnotesize\textbf{2}}
& \multicolumn{1}{c}{\footnotesize\textbf{3}}
& \multicolumn{1}{c}{\footnotesize\textbf{4}}
& \multicolumn{1}{c}{\footnotesize\textbf{5}}
& \\

\cmidrule(r){1-6}

\multicolumn{6}{>{\columncolor[gray]{.95}}l}
{\textit{Static \& Zero-Shot Baselines}}
& \\

Random (Single run)
& 0.479 $\pm$ 0.071 & 0.469 $\pm$ 0.071 & 0.438 $\pm$ 0.057 & 0.388 $\pm$ 0.046 & 0.610 $\pm$ 0.025 & \\

LLM Rec. (Desc.)
& 0.513
& 0.544
& 0.423
& 0.547
& 0.477 & \\

LLM Rec. (Hist.)
& 0.465
& \textbf{0.630}
& 0.435
& 0.558
& 0.461 & \\

Global
& 0.556
& \textit{0.628}
& \textbf{0.632}
& 0.624
& \textit{0.642} & \\

Offline-only (Ranker)
& 0.515 $\pm$ 0.020 & 0.614 $\pm$ 0.008 & 0.613 $\pm$ 0.002 & 0.637 $\pm$ 0.020 & 0.567 $\pm$ 0.005 & \\

\cmidrule(r){1-6}

\multicolumn{6}{>{\columncolor[gray]{.95}}l}
{\textit{Sequential Baselines}}
& \\

Random Search (ES)
& 0.518 $\pm$ 0.056 & 0.554 $\pm$ 0.070 & 0.599 $\pm$ 0.023 & 0.522 $\pm$ 0.021 & 0.610 $\pm$ 0.025 &  \\

Random Search (Full)
& 0.603 $\pm$ 0.028 & 0.617 $\pm$ 0.017 & 0.609 $\pm$ 0.013 & 0.587 $\pm$ 0.020 & 0.617 $\pm$ 0.025 &  \\

BOHB
& \textit{0.620 $\pm$ 0.004} & 0.620 $\pm$ 0.008 & 0.581 $\pm$ 0.021 & 0.584 $\pm$ 0.008 & 0.629 $\pm$ 0.008 &  \\

RGPE (ES)
& 0.456 $\pm$ 0.000 & 0.532 $\pm$ 0.000 & 0.573 $\pm$ 0.043 & 0.527 $\pm$ 0.040 & 0.607 $\pm$ 0.004 &  \\

RGPE (Full)
& \textbf{0.650 $\pm$ 0.016} & 0.622 $\pm$ 0.002 & 0.624 $\pm$ 0.006 & 0.647 $\pm$ 0.009 & 0.638 $\pm$ 0.017 &  \\

SMAC (ES)
& 0.615 $\pm$ 0.003 & 0.620 $\pm$ 0.010 & \textit{0.628 $\pm$ 0.011} & 0.481 $\pm$ 0.077 & 0.583 $\pm$ 0.011 &  \\

SMAC (Full)
& 0.616 $\pm$ 0.015 & 0.626 $\pm$ 0.009 & 0.617 $\pm$ 0.008 & 0.620 $\pm$ 0.006 & 0.618 $\pm$ 0.024 &  \\

TPE (ES)
& 0.556 $\pm$ 0.065 & 0.573 $\pm$ 0.033 & 0.621 $\pm$ 0.016 & 0.568 $\pm$ 0.032 & 0.596 $\pm$ 0.004 &  \\

TPE (Full)
& 0.610 $\pm$ 0.021 & 0.627 $\pm$ 0.008 & 0.613 $\pm$ 0.011 & 0.568 $\pm$ 0.047 & \textbf{0.645 $\pm$ 0.009} &  \\

\cmidrule(r){1-6}

\multicolumn{6}{>{\columncolor[gray]{.95}}l}
{\textit{Ablations}}
& \\

No Residual
& 0.594 $\pm$ 0.021 & 0.616 $\pm$ 0.001 & 0.621 $\pm$ 0.000 & 0.629 $\pm$ 0.026 & 0.588 $\pm$ 0.016 &  \\

No BO
& 0.563 $\pm$ 0.035 & 0.482 $\pm$ 0.135 & 0.618 $\pm$ 0.003 & \textit{0.677 $\pm$ 0.002} & 0.598 $\pm$ 0.017 &  \\

No Offline
& 0.590 $\pm$ 0.015 & 0.581 $\pm$ 0.039 & 0.607 $\pm$ 0.047 & 0.578 $\pm$ 0.033 & 0.515 $\pm$ 0.050 &  \\

\cmidrule(r){1-6}

\multicolumn{6}{>{\columncolor[gray]{.95}}l}
{\textit{Sensitivity Analysis}}
& \\

25\%
& 0.584 $\pm$ 0.018 & 0.613 $\pm$ 0.002 & 0.616 $\pm$ 0.003 & 0.604 $\pm$ 0.007 & 0.588 $\pm$ 0.016 &  \\

75\%
& 0.604 $\pm$ 0.009 & 0.616 $\pm$ 0.001 & 0.616 $\pm$ 0.003 & 0.678 $\pm$ 0.003 & 0.585 $\pm$ 0.018 &  \\

$w_t \ 1.0$ 
& 0.578 $\pm$ 0.043 & 0.619 $\pm$ 0.002 & 0.612 $\pm$ 0.005 & 0.515 $\pm$ 0.161 & 0.588 $\pm$ 0.016 &  \\

$w_t \ 0.7$
& 0.578 $\pm$ 0.043 & 0.619 $\pm$ 0.002 & 0.612 $\pm$ 0.005 & 0.515 $\pm$ 0.161 & 0.588 $\pm$ 0.016 &  \\

$w_t \ 1.0 \to 0.3$ 
& 0.578 $\pm$ 0.043 & 0.619 $\pm$ 0.002 & 0.612 $\pm$ 0.005 & 0.516 $\pm$ 0.161 & 0.588 $\pm$ 0.016 &   \\

\cmidrule[\heavyrulewidth](r){1-6}

Ours 
& 0.562 $\pm$ 0.036 & 0.616 $\pm$ 0.001 & 0.620 $\pm$ 0.001 & \textbf{0.679 $\pm$ 0.002} & 0.619 $\pm$ 0.010 &  \\

\cmidrule[\heavyrulewidth](r){1-6}

\noalign{\vskip 8pt}

\toprule

& \multicolumn{6}{c}{\textbf{Datasets 6--10 and Overall Average}} \\
\cmidrule(lr){2-7}

\textbf{Method}
& \multicolumn{1}{c}{\footnotesize\textbf{6}}
& \multicolumn{1}{c}{\footnotesize\textbf{7}}
& \multicolumn{1}{c}{\footnotesize\textbf{8}}
& \multicolumn{1}{c}{\footnotesize\textbf{9}}
& \multicolumn{1}{c}{\footnotesize\textbf{10}}
& \multicolumn{1}{c}{\footnotesize\textbf{Avg}} \\

\midrule

\multicolumn{7}{>{\columncolor[gray]{.95}}l}
{\textit{Static \& Zero-Shot Baselines}} \\

Random (Single run)
& 0.327 $\pm$ 0.101 & 0.182 $\pm$ 0.092 & 0.280 $\pm$ 0.140 & 0.079 $\pm$ 0.005 & 0.365 $\pm$ 0.106 & 0.362 $\pm$ 0.015 \\

LLM Rec. (Desc.)
& 0.371
& 0.088
& 0.018
& 0.052
& 0.035
& 0.307 \\

LLM Rec. (Hist.)
& 0.382
& 0.532
& 0.410
& 0.431
& 0.117
& 0.442 \\

Global
& 0.622
& 0.454
& \textbf{0.659}
& 0.018
& 0.451
& 0.529 \\

Offline-only (Ranker)
& 0.572 $\pm$ 0.070 & 0.585 $\pm$ 0.023 & 0.584 $\pm$ 0.027 & 0.272 $\pm$ 0.038 & \textit{0.608 $\pm$ 0.023} & 0.557 $\pm$ 0.016 \\

\midrule

\rowcolor[gray]{.95}
\multicolumn{7}{l}{\textit{Sequential Baselines}} \\

Random Search (ES)
& 0.508 $\pm$ 0.007 & 0.515 $\pm$ 0.006 & 0.555 $\pm$ 0.042 & 0.297 $\pm$ 0.048 & 0.423 $\pm$ 0.191 & 0.510 $\pm$ 0.033 \\

Random Search (Full)
& 0.566 $\pm$ 0.031 & 0.515 $\pm$ 0.006 & 0.577 $\pm$ 0.025 & 0.455 $\pm$ 0.034 & 0.584 $\pm$ 0.030 & 0.573 $\pm$ 0.007 \\

BOHB
& \textit{0.628 $\pm$ 0.004} & 0.596 $\pm$ 0.010 & 0.600 $\pm$ 0.015 & 0.472 $\pm$ 0.018 & 0.597 $\pm$ 0.013 & 0.593 $\pm$ 0.002 \\

RGPE (ES)
& 0.573 $\pm$ 0.000 & 0.561 $\pm$ 0.046 & 0.567 $\pm$ 0.021 & 0.412 $\pm$ 0.033 & 0.532 $\pm$ 0.045 & 0.534 $\pm$ 0.013 \\

RGPE (Full)
& \textbf{0.632 $\pm$ 0.006} & \textit{0.617 $\pm$ 0.003} & 0.626 $\pm$ 0.017 & 0.289 $\pm$ 0.015 & \textbf{0.641 $\pm$ 0.010} & \textbf{0.599 $\pm$ 0.002} \\

SMAC (ES)
& 0.524 $\pm$ 0.069 & 0.285 $\pm$ 0.083 & 0.538 $\pm$ 0.057 & 0.282 $\pm$ 0.028 & 0.553 $\pm$ 0.047 & 0.511 $\pm$ 0.003 \\

SMAC (Full)
& 0.615 $\pm$ 0.019 & 0.507 $\pm$ 0.039 & 0.613 $\pm$ 0.021 & \textbf{0.504 $\pm$ 0.019} & 0.567 $\pm$ 0.047 & 0.590 $\pm$ 0.006 \\

TPE (ES)
& 0.501 $\pm$ 0.043 & 0.515 $\pm$ 0.065 & \textit{0.634 $\pm$ 0.017} & 0.251 $\pm$ 0.068 & 0.573 $\pm$ 0.017 & 0.539 $\pm$ 0.006 \\

TPE (Full)
& 0.601 $\pm$ 0.009 & 0.542 $\pm$ 0.005 & 0.621 $\pm$ 0.008 & \textit{0.489 $\pm$ 0.009} & 0.567 $\pm$ 0.050 & 0.588 $\pm$ 0.010 \\

\midrule

\rowcolor[gray]{.95}
\multicolumn{7}{l}{\textit{Ablations}} \\

No Residual
& 0.624 $\pm$ 0.008 & 0.521 $\pm$ 0.101 & 0.587 $\pm$ 0.024 & 0.221 $\pm$ 0.103 & 0.493 $\pm$ 0.131 & 0.549 $\pm$ 0.017 \\

No BO
& 0.615 $\pm$ 0.008 & 0.454 $\pm$ 0.131 & 0.579 $\pm$ 0.020 & 0.382 $\pm$ 0.097 & 0.555 $\pm$ 0.069 & 0.552 $\pm$ 0.019 \\

No Offline
& 0.419 $\pm$ 0.073 & 0.554 $\pm$ 0.017 & 0.602 $\pm$ 0.034 & 0.275 $\pm$ 0.059 & 0.389 $\pm$ 0.092 & 0.511 $\pm$ 0.020 \\

\midrule

\rowcolor[gray]{.95}
\multicolumn{7}{l}{\textit{Sensitivity Analysis}} \\

25\%
& 0.616 $\pm$ 0.014 & 0.542 $\pm$ 0.000 & 0.599 $\pm$ 0.044 & 0.289 $\pm$ 0.007 & 0.561 $\pm$ 0.019 & 0.561 $\pm$ 0.006 \\

75\%
& 0.585 $\pm$ 0.013 & 0.551 $\pm$ 0.021 & 0.596 $\pm$ 0.036 & 0.446 $\pm$ 0.068 & 0.574 $\pm$ 0.012 & 0.585 $\pm$ 0.007 \\

$w_t \ 1.0$
& 0.615 $\pm$ 0.008 & 0.612 $\pm$ 0.006 & 0.579 $\pm$ 0.020 & 0.166 $\pm$ 0.077 & 0.570 $\pm$ 0.014 & 0.545 $\pm$ 0.008 \\

$w_t \ 0.7$
& 0.615 $\pm$ 0.008 & 0.418 $\pm$ 0.209 & 0.579 $\pm$ 0.020 & 0.166 $\pm$ 0.077 & 0.606 $\pm$ 0.024 & 0.530 $\pm$ 0.030 \\

$w_t \ 1.0 \to 0.3$
& 0.615 $\pm$ 0.008 & \textbf{0.620 $\pm$ 0.007} & 0.587 $\pm$ 0.024 & 0.166 $\pm$ 0.077 & 0.601 $\pm$ 0.018 & 0.550 $\pm$ 0.009 \\

\midrule

Ours
& 0.624 $\pm$ 0.008 & 0.612 $\pm$ 0.006 & 0.594 $\pm$ 0.029 & 0.426 $\pm$ 0.074 & 0.588 $\pm$ 0.034 & \textit{0.594 $\pm$ 0.002} \\

\bottomrule
\end{tabularx}

\end{table*}

\begin{table*}[p]
\footnotesize
\setlength{\tabcolsep}{2pt}
\renewcommand{\arraystretch}{0.95}
\caption{Configuration selection regret across ten held-out datasets under a fixed number of online evaluations. Regret is calculated relative to the best-performing configuration in the reference pool for each dataset. Bold indicates the best performance (lowest regret) per dataset; italics indicate the second-best.}
\label{tab:regret-expanded}

\begin{tabularx}{\textwidth}{@{}MYYYYYY@{}}

\cmidrule[\heavyrulewidth](r){1-6}

& \multicolumn{5}{c}{\textbf{Datasets 1--5}} & \\
\cmidrule(lr){2-6}

\textbf{Method}
& \multicolumn{1}{c}{\footnotesize\textbf{1}}
& \multicolumn{1}{c}{\footnotesize\textbf{2}}
& \multicolumn{1}{c}{\footnotesize\textbf{3}}
& \multicolumn{1}{c}{\footnotesize\textbf{4}}
& \multicolumn{1}{c}{\footnotesize\textbf{5}}
& \\

\cmidrule(r){1-6}

\multicolumn{6}{>{\columncolor[gray]{.95}}l}
{\textit{Static \& Zero-Shot Baselines}}
& \\

Random (Single run)
& 0.194 $\pm$ 0.071 & 0.171 $\pm$ 0.071 & 0.226 $\pm$ 0.057 & 0.295 $\pm$ 0.046 & 0.061 $\pm$ 0.025 & \\

LLM Rec. (Desc.) 
& 0.160 & 0.096 & 0.241 & 0.136 & 0.194 & \\

LLM Rec. (Hist.)
& 0.208 & \textbf{0.010} & 0.229 & 0.125 & 0.210 & \\

Global 
& 0.117 & \textit{0.012} & \textbf{0.032} & 0.059 & \textit{0.029} & \\

Offline-only (Ranker)
& 0.158 $\pm$ 0.020 & 0.026 $\pm$ 0.008 & 0.051 $\pm$ 0.002 & 0.046 $\pm$ 0.020 & 0.104 $\pm$ 0.005 & \\

\cmidrule(r){1-6}

\multicolumn{6}{>{\columncolor[gray]{.95}}l}
{\textit{Sequential Baselines}}
& \\

Random Search (ES)
& 0.155 $\pm$ 0.056 & 0.086 $\pm$ 0.070 & 0.065 $\pm$ 0.023 & 0.161 $\pm$ 0.021 & 0.061 $\pm$ 0.025 &  \\

Random Search (Full)
& 0.070 $\pm$ 0.028 & 0.023 $\pm$ 0.017 & 0.055 $\pm$ 0.013 & 0.096 $\pm$ 0.020 & 0.054 $\pm$ 0.025 &  \\

BOHB
& \textit{0.053 $\pm$ 0.004} & 0.020 $\pm$ 0.008 & 0.083 $\pm$ 0.021 & 0.099 $\pm$ 0.008 & 0.042 $\pm$ 0.008 &  \\

RGPE (ES)
& 0.217 $\pm$ 0.000 & 0.108 $\pm$ 0.000 & 0.091 $\pm$ 0.043 & 0.156 $\pm$ 0.040 & 0.064 $\pm$ 0.004 &  \\

RGPE (Full)
& \textbf{0.023 $\pm$ 0.016} & 0.018 $\pm$ 0.002 & 0.040 $\pm$ 0.006 & 0.036 $\pm$ 0.009 & 0.033 $\pm$ 0.017 &  \\

SMAC (ES)
& 0.058 $\pm$ 0.003 & 0.020 $\pm$ 0.010 & \textit{0.036 $\pm$ 0.011} & 0.202 $\pm$ 0.077 & 0.088 $\pm$ 0.011 &  \\

SMAC (Full)
& 0.057 $\pm$ 0.015 & 0.014 $\pm$ 0.009 & 0.047 $\pm$ 0.008 & 0.063 $\pm$ 0.006 & 0.053 $\pm$ 0.024 &  \\

TPE (ES)
& 0.117 $\pm$ 0.065 & 0.067 $\pm$ 0.033 & 0.043 $\pm$ 0.016 & 0.115 $\pm$ 0.032 & 0.075 $\pm$ 0.004 &  \\

TPE (Full)
& 0.063 $\pm$ 0.021 & 0.013 $\pm$ 0.008 & 0.051 $\pm$ 0.011 & 0.115 $\pm$ 0.047 & \textbf{0.026 $\pm$ 0.009} &  \\

\cmidrule(r){1-6}

\multicolumn{6}{>{\columncolor[gray]{.95}}l}
{\textit{Ablations}}
& \\

No Residual
& 0.079 $\pm$ 0.021 & 0.024 $\pm$ 0.001 & 0.043 $\pm$ 0.000 & 0.054 $\pm$ 0.026 & 0.083 $\pm$ 0.016 &  \\

No BO
& 0.110 $\pm$ 0.035 & 0.158 $\pm$ 0.135 & 0.046 $\pm$ 0.003 & \textit{0.006 $\pm$ 0.002} & 0.073 $\pm$ 0.017 &  \\

No Offline
& 0.083 $\pm$ 0.015 & 0.059 $\pm$ 0.039 & 0.057 $\pm$ 0.047 & 0.105 $\pm$ 0.033 & 0.156 $\pm$ 0.050 &  \\

\cmidrule(r){1-6}

\multicolumn{6}{>{\columncolor[gray]{.95}}l}
{\textit{Sensitivity Analysis}}
& \\

25\%
& 0.089 $\pm$ 0.018 & 0.027 $\pm$ 0.002 & 0.048 $\pm$ 0.003 & 0.079 $\pm$ 0.007 & 0.083 $\pm$ 0.016 &  \\

75\%
& 0.069 $\pm$ 0.009 & 0.024 $\pm$ 0.001 & 0.048 $\pm$ 0.003 & 0.005 $\pm$ 0.003 & 0.086 $\pm$ 0.018 &  \\

$w_t \ 1.0$ 
& 0.095 $\pm$ 0.043 & 0.021 $\pm$ 0.002 & 0.052 $\pm$ 0.005 & 0.168 $\pm$ 0.161 & 0.083 $\pm$ 0.016 &  \\

$w_t \ 0.7$
& 0.095 $\pm$ 0.043 & 0.021 $\pm$ 0.002 & 0.052 $\pm$ 0.005 & 0.168 $\pm$ 0.161 & 0.083 $\pm$ 0.016 &  \\

$w_t \ 1.0 \to 0.3$ 
& 0.095 $\pm$ 0.043 & 0.021 $\pm$ 0.002 & 0.052 $\pm$ 0.005 & 0.167 $\pm$ 0.161 & 0.083 $\pm$ 0.016 &   \\

\cmidrule[\heavyrulewidth](r){1-6}

Ours 
& 0.111 $\pm$ 0.036 & 0.024 $\pm$ 0.001 & 0.044 $\pm$ 0.001 & \textbf{0.004 $\pm$ 0.002} & 0.052 $\pm$ 0.010 &  \\

\cmidrule[\heavyrulewidth](r){1-6}

\noalign{\vskip 8pt}

\toprule

& \multicolumn{6}{c}{\textbf{Datasets 6--10 and Overall Average}} \\
\cmidrule(lr){2-7}

\textbf{Method}
& \multicolumn{1}{c}{\footnotesize\textbf{6}}
& \multicolumn{1}{c}{\footnotesize\textbf{7}}
& \multicolumn{1}{c}{\footnotesize\textbf{8}}
& \multicolumn{1}{c}{\footnotesize\textbf{9}}
& \multicolumn{1}{c}{\footnotesize\textbf{10}}
& \multicolumn{1}{c}{\footnotesize\textbf{Avg}} \\

\midrule

\multicolumn{7}{>{\columncolor[gray]{.95}}l}
{\textit{Static \& Zero-Shot Baselines}} \\

Random (Single run)
& 0.322 $\pm$ 0.101 & 0.448 $\pm$ 0.092 & 0.379 $\pm$ 0.140 & 0.495 $\pm$ 0.005 & 0.294 $\pm$ 0.106 & 0.289 $\pm$ 0.015 \\

LLM Rec. (Desc.)
& 0.278 & 0.542 & 0.641 & 0.522 & 0.624 & 0.343 \\

LLM Rec. (Hist.)
& 0.267 & 0.098 & 0.249 & 0.143 & 0.542 & 0.208 \\

Global
& 0.027 & 0.176 & \textbf{0.000} & 0.556 & 0.208 & 0.122 \\

Offline-only (Ranker)
& 0.077 $\pm$ 0.070 & 0.045 $\pm$ 0.023 & 0.075 $\pm$ 0.027 & 0.302 $\pm$ 0.038 & \textit{0.051 $\pm$ 0.023} & 0.094 $\pm$ 0.016 \\

\midrule

\rowcolor[gray]{.95}
\multicolumn{7}{l}{\textit{Sequential Baselines}} \\

Random Search (ES)
& 0.141 $\pm$ 0.007 & 0.115 $\pm$ 0.006 & 0.104 $\pm$ 0.042 & 0.277 $\pm$ 0.048 & 0.236 $\pm$ 0.191 & 0.140 $\pm$ 0.033 \\

Random Search (Full)
& 0.083 $\pm$ 0.031 & 0.115 $\pm$ 0.006 & 0.082 $\pm$ 0.025 & 0.119 $\pm$ 0.034 & 0.075 $\pm$ 0.030 & 0.077 $\pm$ 0.007 \\

BOHB
& \textit{0.021 $\pm$ 0.004} & 0.034 $\pm$ 0.010 & 0.059 $\pm$ 0.015 & 0.102 $\pm$ 0.018 & 0.062 $\pm$ 0.013 & 0.057 $\pm$ 0.002 \\

RGPE (ES)
& 0.076 $\pm$ 0.000 & 0.069 $\pm$ 0.046 & 0.092 $\pm$ 0.021 & 0.162 $\pm$ 0.033 & 0.127 $\pm$ 0.045 & 0.116 $\pm$ 0.013 \\

RGPE (Full)
& \textbf{0.017 $\pm$ 0.006} & \textit{0.013 $\pm$ 0.003} & 0.033 $\pm$ 0.017 & 0.285 $\pm$ 0.015 & \textbf{0.018 $\pm$ 0.010} & \textbf{0.052 $\pm$ 0.002} \\

SMAC (ES)
& 0.125 $\pm$ 0.069 & 0.345 $\pm$ 0.083 & 0.121 $\pm$ 0.057 & 0.292 $\pm$ 0.028 & 0.106 $\pm$ 0.047 & 0.139 $\pm$ 0.003 \\

SMAC (Full)
& 0.034 $\pm$ 0.019 & 0.123 $\pm$ 0.039 & 0.046 $\pm$ 0.021 & \textbf{0.070 $\pm$ 0.019} & 0.092 $\pm$ 0.047 & 0.060 $\pm$ 0.006 \\

TPE (ES)
& 0.148 $\pm$ 0.043 & 0.115 $\pm$ 0.065 & \textit{0.025 $\pm$ 0.017} & 0.323 $\pm$ 0.068 & 0.086 $\pm$ 0.017 & 0.111 $\pm$ 0.006 \\

TPE (Full)
& 0.048 $\pm$ 0.009 & 0.088 $\pm$ 0.005 & 0.038 $\pm$ 0.008 & \textit{0.085 $\pm$ 0.009} & 0.092 $\pm$ 0.050 & 0.062 $\pm$ 0.010 \\

\midrule

\rowcolor[gray]{.95}
\multicolumn{7}{l}{\textit{Ablations}} \\

No Residual
& 0.025 $\pm$ 0.008 & 0.109 $\pm$ 0.101 & 0.072 $\pm$ 0.024 & 0.353 $\pm$ 0.103 & 0.166 $\pm$ 0.131 & 0.101 $\pm$ 0.017 \\

No BO
& 0.034 $\pm$ 0.008 & 0.176 $\pm$ 0.131 & 0.080 $\pm$ 0.020 & 0.192 $\pm$ 0.097 & 0.104 $\pm$ 0.069 & 0.098 $\pm$ 0.019 \\

No Offline
& 0.230 $\pm$ 0.073 & 0.076 $\pm$ 0.017 & 0.057 $\pm$ 0.034 & 0.299 $\pm$ 0.059 & 0.270 $\pm$ 0.092 & 0.139 $\pm$ 0.020 \\

\midrule

\rowcolor[gray]{.95}
\multicolumn{7}{l}{\textit{Sensitivity Analysis}} \\

25\%
& 0.033 $\pm$ 0.014 & 0.088 $\pm$ 0.000 & 0.060 $\pm$ 0.044 & 0.285 $\pm$ 0.007 & 0.098 $\pm$ 0.019 & 0.089 $\pm$ 0.006 \\

75\%
& 0.064 $\pm$ 0.013 & 0.079 $\pm$ 0.021 & 0.063 $\pm$ 0.036 & 0.128 $\pm$ 0.068 & 0.085 $\pm$ 0.012 & 0.065 $\pm$ 0.007 \\

$w_t \ 1.0$
& 0.034 $\pm$ 0.008 & 0.018 $\pm$ 0.006 & 0.080 $\pm$ 0.020 & 0.408 $\pm$ 0.077 & 0.089 $\pm$ 0.014 & 0.105 $\pm$ 0.008 \\

$w_t \ 0.7$
& 0.034 $\pm$ 0.008 & 0.212 $\pm$ 0.209 & 0.080 $\pm$ 0.020 & 0.408 $\pm$ 0.077 & 0.053 $\pm$ 0.024 & 0.121 $\pm$ 0.030 \\

$w_t \ 1.0 \to 0.3$
& 0.034 $\pm$ 0.008 & \textbf{0.010 $\pm$ 0.007} & 0.072 $\pm$ 0.024 & 0.408 $\pm$ 0.077 & 0.058 $\pm$ 0.018 & 0.100 $\pm$ 0.009 \\

\midrule

Ours
& 0.025 $\pm$ 0.008 & 0.018 $\pm$ 0.006 & 0.065 $\pm$ 0.029 & 0.148 $\pm$ 0.074 & 0.071 $\pm$ 0.034 & \textit{0.056 $\pm$ 0.002} \\

\bottomrule
\end{tabularx}

\end{table*}

\begin{table*}[p]
\footnotesize
\setlength{\tabcolsep}{2pt}
\renewcommand{\arraystretch}{0.95}
\caption{Configuration selection cost across ten held-out datasets under a fixed number of online evaluations. Cost is expressed in terms of H200 hours.}
\label{tab:cost-expanded}

\begin{tabularx}{\textwidth}{@{}MYYYYYY@{}}

\cmidrule[\heavyrulewidth](r){1-6}

& \multicolumn{5}{c}{\textbf{Datasets 1--5}} & \\
\cmidrule(lr){2-6}

\textbf{Method}
& \multicolumn{1}{c}{\footnotesize\textbf{1}}
& \multicolumn{1}{c}{\footnotesize\textbf{2}}
& \multicolumn{1}{c}{\footnotesize\textbf{3}}
& \multicolumn{1}{c}{\footnotesize\textbf{4}}
& \multicolumn{1}{c}{\footnotesize\textbf{5}}
& \\

\cmidrule(r){1-6}

\multicolumn{6}{>{\columncolor[gray]{.95}}l}
{\textit{Static \& Zero-Shot Baselines}}
& \\

Random (Single run)
& 2.8 $\pm$ 0.7 & 2.5 $\pm$ 0.3 & 4.5 $\pm$ 0.8 & 2.5 $\pm$ 0.9 & 9.8 $\pm$ 1.5 & \\

LLM Rec. (Desc.)
& 8.0 & 8.6 & 8.7 & 11.6 & 10.3 & \\

LLM Rec. (Hist.)
& 4.1 & 5.3 & 4.2 & 6.6 & 11.5 & \\

Global
& 3.6 & 6.1 & 3.8 & 4.7 & 4.5 & \\

Offline-only (Ranker)
& 7.7 $\pm$ 0.6 & 5.1 $\pm$ 0.2 & 5.6 $\pm$ 0.2 & 7.8 $\pm$ 0.5 & 10.4 $\pm$ 0.6 & \\

\cmidrule(r){1-6}

\multicolumn{6}{>{\columncolor[gray]{.95}}l}
{\textit{Sequential Baselines}}
& \\

Random Search (ES)
& 4.6 $\pm$ 0.7 & 4.6 $\pm$ 1.6 & 5.3 $\pm$ 0.9 & 5.7 $\pm$ 1.1 & 11 $\pm$ 1.6 &  \\

Random Search (Full)
& 57.4 $\pm$ 2.0 & 72.5 $\pm$ 3.7 & 60.7 $\pm$ 5.1 & 72.1 $\pm$ 8.2 & 81.9 $\pm$ 5.1 &  \\

BOHB
& 100.8 $\pm$ 1.1 & 119.0 $\pm$ 13.7 & 110.6 $\pm$ 11.5 & 129.7 $\pm$ 13.4 & 131.5 $\pm$ 11.8 &  \\

RGPE (ES)
& 10.6 $\pm$ 0.2 & 8.8 $\pm$ 0.1 & 8.9 $\pm$ 0.9 & 12.2 $\pm$ 0.7 & 7.5 $\pm$ 0.5 &  \\

RGPE (Full)
& 76.6 $\pm$ 4.8 & 52.0 $\pm$ 5.0 & 69.5 $\pm$ 5.0 & 111.1 $\pm$ 2.7 & 86.2 $\pm$ 2.3 &  \\

SMAC (ES)
& 5.0 $\pm$ 0.2 & 6.0 $\pm$ 1.7 & 7.6 $\pm$ 1.7 & 6.4 $\pm$ 0.5 & 10.3 $\pm$ 1.8 &  \\

SMAC (Full)
& 76.3 $\pm$ 2.2 & 55.8 $\pm$ 5.2 & 65.2 $\pm$ 2.7 & 98.0 $\pm$ 8.3 & 111.6 $\pm$ 6.8 &  \\

TPE (ES)
& 4.8 $\pm$ 0.3 & 6.0 $\pm$ 1.4 & 5.5 $\pm$ 0.3 & 6.1 $\pm$ 0.3 & 11.7 $\pm$ 2.0 &  \\

TPE (Full)
& 61.1 $\pm$ 3.0 & 73.8 $\pm$ 1.1 & 77.8 $\pm$ 2.3 & 74.9 $\pm$ 11.1 & 72.0 $\pm$ 9.5 &  \\

\cmidrule(r){1-6}

\multicolumn{6}{>{\columncolor[gray]{.95}}l}
{\textit{Ablations}}
& \\

No Residual
& 10.6 $\pm$ 0.3 & 7.2 $\pm$ 0.4 & 7.8 $\pm$ 0.9 & 10.6 $\pm$ 0.7 & 13.1 $\pm$ 0.6 &  \\

No BO
& 12.6 $\pm$ 0.9 & 9.1 $\pm$ 1.5 & 8.6 $\pm$ 0.7 & 11.8 $\pm$ 0.3 & 12.4 $\pm$ 1.5 &  \\

No Offline
& 8.0 $\pm$ 0.5 & 3.9 $\pm$ 0.1 & 6.8 $\pm$ 1.5 & 12.8 $\pm$ 2.1 & 10.0 $\pm$ 2.1 &  \\

\cmidrule(r){1-6}

\multicolumn{6}{>{\columncolor[gray]{.95}}l}
{\textit{Sensitivity Analysis}}
& \\

25\%
& 11.1 $\pm$ 0.6 & 6.8 $\pm$ 0.7 & 8.2 $\pm$ 0.8 & 10.1 $\pm$ 0.8 & 13.0 $\pm$ 0.9 &  \\

75\%
& 11.1 $\pm$ 0.4 & 7.6 $\pm$ 0.1 & 8.7 $\pm$ 0.4 & 11.7 $\pm$ 0.1 & 14.2 $\pm$ 0.3 &  \\

$w_t \ 1.0$ 
& 11.8 $\pm$ 0.7 & 7.8 $\pm$ 0.1 & 8.6 $\pm$ 0.4 & 11.7 $\pm$ 0.3 & 13.4 $\pm$ 0.7 &  \\

$w_t \ 0.7$
& 11.8 $\pm$ 0.8 & 7.8 $\pm$ 0.1 & 8.5 $\pm$ 0.5 & 11.7 $\pm$ 0.2 & 13.4 $\pm$ 0.6 &  \\

$w_t \ 1.0 \to 0.3$ 
& 11.8 $\pm$ 0.8 & 7.8 $\pm$ 0.1 & 9.2 $\pm$ 0.8 & 12.5 $\pm$ 0.6 & 13.5 $\pm$ 0.6 &   \\

\cmidrule[\heavyrulewidth](r){1-6}

Ours 
& 11.5 $\pm$ 0.9 & 7.5 $\pm$ 0.4 & 8.0 $\pm$ 0.7 & 11.7 $\pm$ 0.4 & 13.1 $\pm$ 0.9 &  \\

\cmidrule[\heavyrulewidth](r){1-6}

\noalign{\vskip 8pt}

\toprule

& \multicolumn{6}{c}{\textbf{Datasets 6--10 and Overall Average}} \\
\cmidrule(lr){2-7}

\textbf{Method}
& \multicolumn{1}{c}{\footnotesize\textbf{6}}
& \multicolumn{1}{c}{\footnotesize\textbf{7}}
& \multicolumn{1}{c}{\footnotesize\textbf{8}}
& \multicolumn{1}{c}{\footnotesize\textbf{9}}
& \multicolumn{1}{c}{\footnotesize\textbf{10}}
& \multicolumn{1}{c}{\footnotesize\textbf{Avg}} \\

\midrule

\multicolumn{7}{>{\columncolor[gray]{.95}}l}
{\textit{Static \& Zero-Shot Baselines}} \\

Random (Single run)
& 3.9 $\pm$ 2.6 & 2.9 $\pm$ 0.9 & 3.7 $\pm$ 0.7 & 5.2 $\pm$ 0.8 & 6.9 $\pm$ 0.8 & 4.5 $\pm$ 0.4 \\

LLM Rec. (Desc.)
& 3.2 & 5.5 & 7.4 & 5 & 7.7 & 7.6 \\

LLM Rec. (Hist.)
& 3.1 & 5.4 & 2.9 & 3.8 & 9.1 & 5.6 \\

Global
& 3.8 & 3.6 & 2.8 & 3.5 & 4.2 & 4.1 \\

Offline-only (Ranker)
& 3.5 $\pm$ 0.0 & 5.7 $\pm$ 1.1 & 5.3 $\pm$ 0.7 & 6.0 $\pm$ 1.1 & 7.0 $\pm$ 0.8 & 6.4 $\pm$ 0.2 \\

\midrule

\rowcolor[gray]{.95}
\multicolumn{7}{l}{\textit{Sequential Baselines}} \\

Random Search (ES)
& 3.2 $\pm$ 0.4 & 8.1 $\pm$ 2.1 & 5.0 $\pm$ 1.6 & 6.9 $\pm$ 1.7 & 7.3 $\pm$ 1.6 & 6.2 $\pm$ 0.6 \\

Random Search (Full)
& 41.3 $\pm$ 4.5 & 51.0 $\pm$ 2.8 & 49.9 $\pm$ 5.7 & 56.6 $\pm$ 6.2 & 74.1 $\pm$ 9.1 & 61.7 $\pm$ 1.7 \\

BOHB
& 60.1 $\pm$ 10.1 & 76.0 $\pm$ 11.8 & 72.7 $\pm$ 5.4 & 67.7 $\pm$ 7.5 & 117.3 $\pm$ 13.3 & 98.5 $\pm$ 2.1 \\

RGPE (ES)
& 6.5 $\pm$ 1.5 & 6.1 $\pm$ 0.6 & 8.8 $\pm$ 1.8 & 10.7 $\pm$ 1.0 & 9.8 $\pm$ 0.9 & 9.0 $\pm$ 0.2 \\

RGPE (Full)
& 53.7 $\pm$ 3.7 & 53.7 $\pm$ 2.4 & 62.3 $\pm$ 4.8 & 88.9 $\pm$ 2.0 & 102.6 $\pm$ 14 & 75.6 $\pm$ 2.1 \\

SMAC (ES)
& 3.5 $\pm$ 0.7 & 5.0 $\pm$ 1.3 & 5.2 $\pm$ 1.3 & 5.7 $\pm$ 0.2 & 5.3 $\pm$ 0.6 & 6.0 $\pm$ 0.1 \\

SMAC (Full)
& 37.2 $\pm$ 3.1 & 42.2 $\pm$ 2.9 & 41.4 $\pm$ 4.8 & 48.2 $\pm$ 7.2 & 81.2 $\pm$ 8.3 & 65.7 $\pm$ 1.1 \\

TPE (ES)
& 3.2 $\pm$ 0.3 & 4.1 $\pm$ 0.6 & 5.2 $\pm$ 0.3 & 6.9 $\pm$ 1.0 & 6.7 $\pm$ 1.1 & 6.0 $\pm$ 0.2 \\

TPE (Full)
& 58.7 $\pm$ 8.6 & 51.9 $\pm$ 5.4 & 55.7 $\pm$ 3.4 & 52.2 $\pm$ 2.4 & 73.8 $\pm$ 5.5 & 65.2 $\pm$ 0.8 \\

\midrule

\rowcolor[gray]{.95}
\multicolumn{7}{l}{\textit{Ablations}} \\

No Residual
& 6.7 $\pm$ 1.3 & 7.2 $\pm$ 0.2 & 7.4 $\pm$ 0.8 & 8.9 $\pm$ 0.9 & 10.1 $\pm$ 1.2 & 9.0 $\pm$ 0.5 \\

No BO
& 7.5 $\pm$ 0.9 & 7.9 $\pm$ 0.9 & 7.3 $\pm$ 1.3 & 10.0 $\pm$ 0.7 & 11.9 $\pm$ 1.4 & 9.9 $\pm$ 0.4 \\

No Offline
& 5.7 $\pm$ 1.0 & 6.1 $\pm$ 0.9 & 5.3 $\pm$ 0.8 & 6.5 $\pm$ 1.6 & 6.3 $\pm$ 1.4 & 7.1 $\pm$ 0.8 \\

\midrule

\rowcolor[gray]{.95}
\multicolumn{7}{l}{\textit{Sensitivity Analysis}} \\

25\%
& 5.0 $\pm$ 0.7 & 7.5 $\pm$ 0.6 & 8.7 $\pm$ 1.8 & 7.8 $\pm$ 1.2 & 9.1 $\pm$ 0.9 & 8.7 $\pm$ 0.9 \\

75\%
& 7.8 $\pm$ 1.1 & 8.4 $\pm$ 0.1 & 8.0 $\pm$ 1.0 & 10.6 $\pm$ 0.7 & 11.3 $\pm$ 1.6 & 10.0 $\pm$ 0.3 \\

$w_t \ 1.0$
& 6.7 $\pm$ 1.3 & 8.5 $\pm$ 0.9 & 7.2 $\pm$ 1.2 & 9.2 $\pm$ 1.3 & 11.4 $\pm$ 1.3 & 9.6 $\pm$ 0.4 \\

$w_t \ 0.7$
& 6.7 $\pm$ 1.3 & 8.0 $\pm$ 0.5 & 7.6 $\pm$ 0.8 & 9.1 $\pm$ 1.3 & 10.4 $\pm$ 1.2 & 9.5 $\pm$ 0.4 \\

$w_t \ 1.0 \to 0.3$
& 6.6 $\pm$ 1.3 & 9.0 $\pm$ 1.0 & 8.5 $\pm$ 0.4 & 9.1 $\pm$ 1.2 & 10.6 $\pm$ 1.4 & 9.9 $\pm$ 0.1 \\

\midrule

Ours
& 6.8 $\pm$ 1.1 & 7.6 $\pm$ 0.4 & 7.6 $\pm$ 0.7 & 9.2 $\pm$ 0.1 & 11.9 $\pm$ 0.6 & 9.5 $\pm$ 0.4 \\

\bottomrule
\end{tabularx}

\end{table*}

\subsection{Offline Component Analysis}
\label{appendix:results-offline}

Table~\ref{tab:offline-expanded} reports dataset-level offline surrogate ranking performance (in terms of ranking metrics such as regret, recall@k, nDCG, pairwise accuracy and Spearman $\rho$) for both the offline ranker and the early stop regressor. The different ranking diagnostics reveal a difference between improving ranking as a whole and identifying the optimum. Relative to the offline ranker, the early-stop regressor improves more holistic ranking metrics such as nDCG, pairwise accuracy and Spearman correlation, however recall@$k$ decreases. This supports using the early-stop regressor to provide noisy corrective evidence rather than treating it as a final configuration selector.

\begin{table*}[p]
\footnotesize
\setlength{\tabcolsep}{2pt}
\renewcommand{\arraystretch}{0.95}
\caption{Offline surrogate ranking performance across ten held-out datasets under LODO evaluation.}
\label{tab:offline-expanded}

\begin{tabularx}{\textwidth}{@{}MYYYYYY@{}}

\cmidrule[\heavyrulewidth](r){1-6}

& \multicolumn{5}{c}{\textbf{Datasets 1--5}} & \\
\cmidrule(lr){2-6}

\textbf{Metric}
& \multicolumn{1}{c}{\footnotesize\textbf{1}}
& \multicolumn{1}{c}{\footnotesize\textbf{2}}
& \multicolumn{1}{c}{\footnotesize\textbf{3}}
& \multicolumn{1}{c}{\footnotesize\textbf{4}}
& \multicolumn{1}{c}{\footnotesize\textbf{5}}
& \\

\cmidrule(r){1-6}

\multicolumn{6}{>{\columncolor[gray]{.95}}l}
{\textit{Offline Ranker}}
& \\

Regret
& 0.075 $\pm$ 0.020 & 0.065 $\pm$ 0.008 & 0.048 $\pm$ 0.000 & 0.027 $\pm$ 0.006 & 0.032 $\pm$ 0.000 &  \\

Recall@1
& 0.000 $\pm$ 0.000 & 0.000 $\pm$ 0.000 & 0.000 $\pm$ 0.000 & 0.000 $\pm$ 0.000 & 1.000 $\pm$ 0.000 & \\

Recall@3
& 0.000 $\pm$ 0.000 & 1.000 $\pm$ 0.000 & 0.000 $\pm$ 0.000 & 0.000 $\pm$ 0.000 & 1.000 $\pm$ 0.000 &   \\

Recall@5
& 0.333 $\pm$ 0.333 & 1.000 $\pm$ 0.000 & 0.667 $\pm$ 0.333 & 0.333 $\pm$ 0.333 & 1.000 $\pm$ 0.000 &   \\

Recall@10
& 0.333 $\pm$ 0.333 & 1.000 $\pm$ 0.000 & 0.667 $\pm$ 0.333 & 0.333 $\pm$ 0.333 & 1.000 $\pm$ 0.000 &  \\

nDCG@5
& 0.739 $\pm$ 0.005 & 0.956 $\pm$ 0.004 & 0.942 $\pm$ 0.001 & 0.973 $\pm$ 0.004 & 0.968 $\pm$ 0.010 &    \\

nDCG@10
& 0.815 $\pm$ 0.005 & 0.918 $\pm$ 0.022 & 0.955 $\pm$ 0.004 & 0.972 $\pm$ 0.007 & 0.972 $\pm$ 0.003 &    \\

Pairwise Accuracy
& 0.785 $\pm$ 0.009 & 0.720 $\pm$ 0.003 & 0.808 $\pm$ 0.004 & 0.798 $\pm$ 0.004 & 0.773 $\pm$ 0.002 &    \\

Spearman $\rho$
& 0.743 $\pm$ 0.015 & 0.607 $\pm$ 0.008 & 0.817 $\pm$ 0.009 & 0.783 $\pm$ 0.010 & 0.728 $\pm$ 0.005 &   \\

\cmidrule(r){1-6}

\multicolumn{6}{>{\columncolor[gray]{.95}}l}
{\textit{Early Stop Regressor}}
& \\

Regret
& 0.060 $\pm$ 0.000 & 0.057 $\pm$ 0.000 & 0.048 $\pm$ 0.000 & 0.021 $\pm$ 0.000 & 0.032 $\pm$ 0.000 &  \\

Recall@1
& 0.000 $\pm$ 0.000 & 0.000 $\pm$ 0.000 & 0.000 $\pm$ 0.000 & 0.000 $\pm$ 0.000 & 0.000 $\pm$ 0.000 &  \\

Recall@3
& 0.000 $\pm$ 0.000 & 0.000 $\pm$ 0.000 & 0.000 $\pm$ 0.000 & 0.000 $\pm$ 0.000 & 0.667 $\pm$ 0.333 &  \\

Recall@5
& 0.000 $\pm$ 0.000 & 0.000 $\pm$ 0.000 & 0.000 $\pm$ 0.000 & 0.000 $\pm$ 0.000 & 1.000 $\pm$ 0.000 &  \\

Recall@10
& 0.000 $\pm$ 0.000 & 1.000 $\pm$ 0.000 & 0.667 $\pm$ 0.333 & 0.000 $\pm$ 0.000 & 1.000 $\pm$ 0.000 &  \\

nDCG@5
& 0.727 $\pm$ 0.003 & 0.918 $\pm$ 0.000 & 0.941 $\pm$ 0.001 & 0.977 $\pm$ 0.001 & 0.979 $\pm$ 0.002 &  \\

nDCG@10
& 0.799 $\pm$ 0.001 & 0.925 $\pm$ 0.001 & 0.956 $\pm$ 0.006 & 0.937 $\pm$ 0.011 & 0.957 $\pm$ 0.002 &  \\

Pairwise Accuracy
& 0.780 $\pm$ 0.001 & 0.750 $\pm$ 0.001 & 0.832 $\pm$ 0.001 & 0.795 $\pm$ 0.001 & 0.792 $\pm$ 0.002 &  \\

Spearman $\rho$
& 0.722 $\pm$ 0.001 & 0.683 $\pm$ 0.001 & 0.848 $\pm$ 0.002 & 0.769 $\pm$ 0.001 & 0.771 $\pm$ 0.003 & \\

\cmidrule[\heavyrulewidth](r){1-6}

\noalign{\vskip 8pt}

\toprule

& \multicolumn{6}{c}{\textbf{Datasets 6--10 and Overall Average}} \\
\cmidrule(lr){2-7}

\textbf{Metric}
& \multicolumn{1}{c}{\footnotesize\textbf{6}}
& \multicolumn{1}{c}{\footnotesize\textbf{7}}
& \multicolumn{1}{c}{\footnotesize\textbf{8}}
& \multicolumn{1}{c}{\footnotesize\textbf{9}}
& \multicolumn{1}{c}{\footnotesize\textbf{10}}
& \multicolumn{1}{c}{\footnotesize\textbf{Avg}} \\

\midrule

\multicolumn{7}{>{\columncolor[gray]{.95}}l}
{\textit{Offline Ranker}} \\

Regret
& 0.213 $\pm$ 0.086 & 0.470 $\pm$ 0.115 & 0.071 $\pm$ 0.001 & 0.000 $\pm$ 0.000 & 0.183 $\pm$ 0.000 & 0.118 $\pm$ 0.013 \\

Recall@1
& 0.000 $\pm$ 0.000 & 0.000 $\pm$ 0.000 & 0.000 $\pm$ 0.000 & 1.000 $\pm$ 0.000 & 0.000 $\pm$ 0.000 & 0.100 $\pm$ 0.000 \\

Recall@3
& 0.667 $\pm$ 0.333 & 0.333 $\pm$ 0.333 & 0.000 $\pm$ 0.000 & 1.000 $\pm$ 0.000 & 0.000 $\pm$ 0.000 & 0.300 $\pm$ 0.058 \\

Recall@5
& 0.667 $\pm$ 0.333 & 0.333 $\pm$ 0.333 & 0.000 $\pm$ 0.000 & 1.000 $\pm$ 0.000 & 0.000 $\pm$ 0.000 & 0.400 $\pm$ 0.058 \\

Recall@10
& 1.000 $\pm$ 0.000 & 0.667 $\pm$ 0.333 & 0.333 $\pm$ 0.333 & 1.000 $\pm$ 0.000 & 0.000 $\pm$ 0.000 & 0.633 $\pm$ 0.033 \\

nDCG@5
& 0.829 $\pm$ 0.018 & 0.583 $\pm$ 0.021 & 0.906 $\pm$ 0.017 & 0.685 $\pm$ 0.012 & 0.822 $\pm$ 0.014 & 0.840 $\pm$ 0.007 \\

nDCG@10
& 0.879 $\pm$ 0.015 & 0.688 $\pm$ 0.017 & 0.938 $\pm$ 0.009 & 0.609 $\pm$ 0.019 & 0.794 $\pm$ 0.039 & 0.854 $\pm$ 0.009 \\

Pairwise Accuracy
& 0.738 $\pm$ 0.004 & 0.716 $\pm$ 0.005 & 0.763 $\pm$ 0.005 & 0.676 $\pm$ 0.005 & 0.692 $\pm$ 0.008 & 0.747 $\pm$ 0.002 \\

Spearman $\rho$
& 0.655 $\pm$ 0.007 & 0.625 $\pm$ 0.006 & 0.729 $\pm$ 0.007 & 0.521 $\pm$ 0.012 & 0.545 $\pm$ 0.023 & 0.675 $\pm$ 0.005 \\

\midrule

\rowcolor[gray]{.95}
\multicolumn{7}{l}{\textit{Early Stop Regressor}} \\

Regret
& 0.299 $\pm$ 0.000 & 0.210 $\pm$ 0.188 & 0.053 $\pm$ 0.017 & 0.000 $\pm$ 0.000 & 0.183 $\pm$ 0.000 & 0.096 $\pm$ 0.017 \\

Recall@1
& 0.000 $\pm$ 0.000 & 0.000 $\pm$ 0.000 & 0.000 $\pm$ 0.000 & 1.000 $\pm$ 0.000 & 0.000 $\pm$ 0.000 & 0.100 $\pm$ 0.000 \\

Recall@3
& 0.667 $\pm$ 0.333 & 0.000 $\pm$ 0.000 & 0.000 $\pm$ 0.000 & 1.000 $\pm$ 0.000 & 0.000 $\pm$ 0.000 & 0.233 $\pm$ 0.067 \\

Recall@5
& 1.000 $\pm$ 0.000 & 0.000 $\pm$ 0.000 & 0.000 $\pm$ 0.000 & 1.000 $\pm$ 0.000 & 0.000 $\pm$ 0.000 & 0.300 $\pm$ 0.000 \\

Recall@10
& 1.000 $\pm$ 0.000 & 0.000 $\pm$ 0.000 & 0.000 $\pm$ 0.000 & 1.000 $\pm$ 0.000 & 0.000 $\pm$ 0.000 & 0.467 $\pm$ 0.033 \\

nDCG@5
& 0.794 $\pm$ 0.002 & 0.652 $\pm$ 0.038 & 0.943 $\pm$ 0.003 & 0.743 $\pm$ 0.034 & 0.837 $\pm$ 0.001 & 0.851 $\pm$ 0.005 \\

nDCG@10
& 0.849 $\pm$ 0.008 & 0.749 $\pm$ 0.028 & 0.880 $\pm$ 0.004 & 0.666 $\pm$ 0.005 & 0.846 $\pm$ 0.014 & 0.856 $\pm$ 0.006 \\

Pairwise Accuracy
& 0.766 $\pm$ 0.000 & 0.755 $\pm$ 0.002 & 0.737 $\pm$ 0.002 & 0.705 $\pm$ 0.002 & 0.723 $\pm$ 0.001 & 0.764 $\pm$ 0.000 \\

Spearman $\rho$
& 0.714 $\pm$ 0.001 & 0.692 $\pm$ 0.005 & 0.666 $\pm$ 0.006 & 0.576 $\pm$ 0.003 & 0.629 $\pm$ 0.004 & 0.707 $\pm$ 0.001 \\

\bottomrule
\end{tabularx}

\end{table*}

\subsection{Sensitivity to Budget}
\label{appendix:results-sensitivity}

\begin{figure*}[p]
\centering
\includegraphics[width=0.9\textwidth]{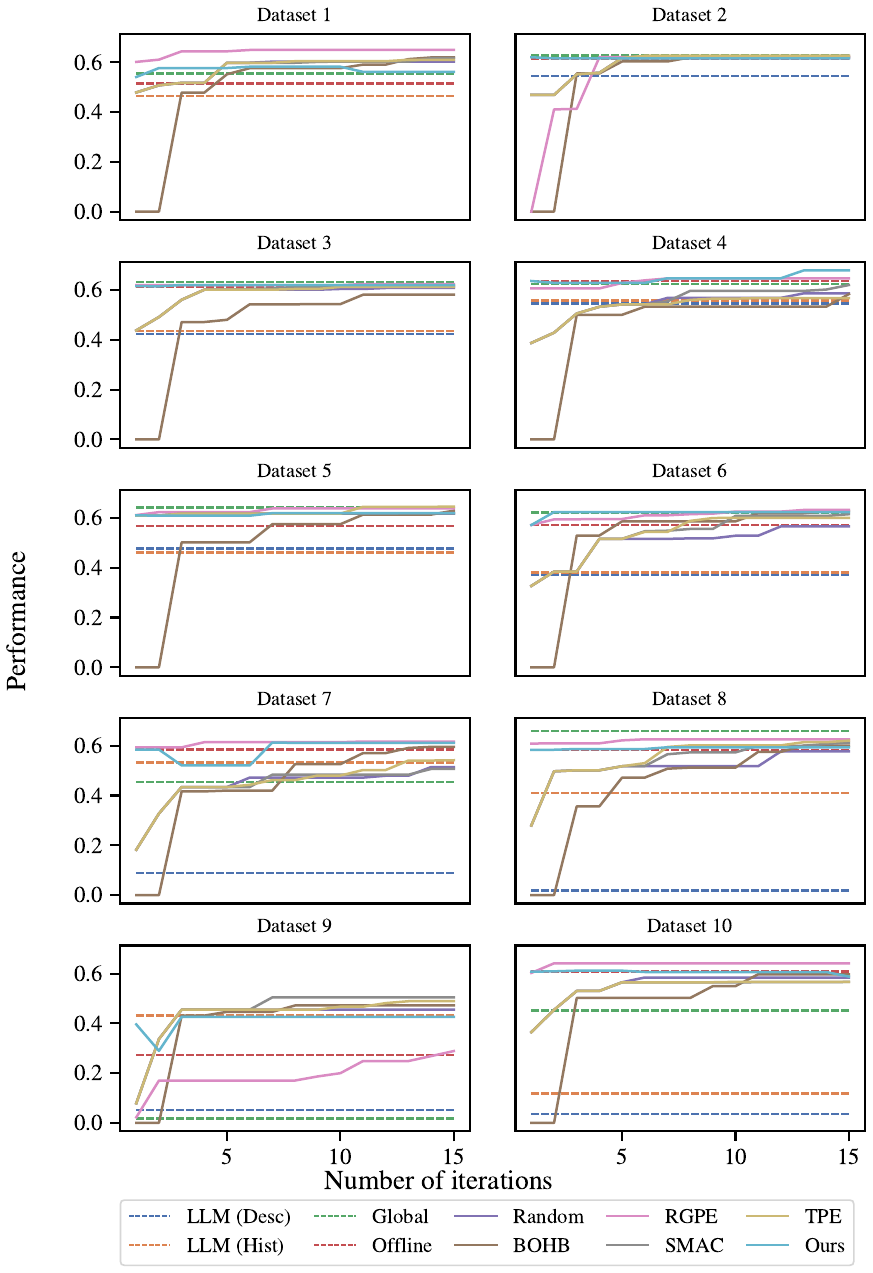}
\caption{Sensitivity of configuration selection performance to the online evaluation budget across ten held-out datasets.}
\label{fig:budget-curves-by-dataset}
\end{figure*}

Figure~\ref{fig:budget-curves-by-dataset} shows how the performance of the best observed configuration varies as the online evaluation budget increases. Across datasets, AutoPipe generally reaches competitive performance within a small number of online evaluations. However, this effect is dataset-dependent; on dataset folds where the initial recommendation is already strong, AutoPipe changes little as budget increases, whereas on other folds online observations produce a visible improvement over the initial recommendation. Thus, increasing evaluations becomes more valuable under larger target-specific mismatch.

\FloatBarrier

\subsection{Source-Disjoint Evaluation on CasiMedicos-Arg}
\label{appendix:casimedicos}

\begin{figure}[h]
\centering
\includegraphics[width=0.9\columnwidth]{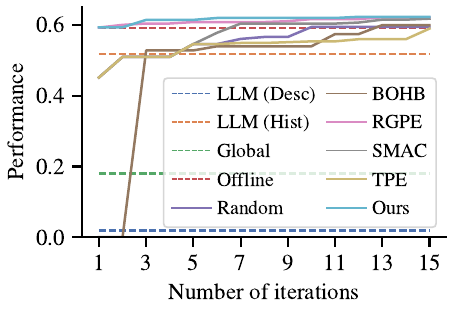}
\caption{Sensitivity of performance (average benchmark accuracy) to online evaluation budget.} 
\label{fig:budget-curves-11}
\end{figure}

We provide extended results for the source-disjoint target introduced in Section~\ref{sec:source-disjoint-transfer}. While the comparisons with other baselines are in Table~\ref{tab:casicomedicos}, we also repeated the component ablations and sensitivity analyses in a source-disjoint setting to determine whether the qualitative behavior observed under the primary LODO evaluation persists. Table~\ref{tab:casicomedicos-extra} and Figure~\ref{fig:budget-curves-11} present our results. 

\begin{table}[h]
\centering
\small
\caption{Results for ablations, sensitivity to early-stop truncation length and sensitivity to offline-guidance tempering in the source-disjoint setting.} 
\label{tab:casicomedicos-extra}
\resizebox{\columnwidth}{!}{%
\begin{tabular}{lccc}
\toprule
\textbf{Method} & \textbf{Performance} & \textbf{Regret} & \textbf{Cost} \\
\midrule
\rowcolor[gray]{.95} \multicolumn{4}{l}{\textit{Ablations}} \\
No Residual & 0.613 $ \pm $ 0.008 & 0.014 $ \pm $ 0.008 & 1.2 $\pm$ 0.1 \\
No BO & 0.614 $ \pm $ 0.008 & 0.013 $ \pm $ 0.008 & 1.3 $\pm$ 0.2 \\
No Offline & 0.585 $ \pm $ 0.031 & 0.042 $ \pm $ 0.031 & 0.7 $\pm$ 0.0 \\
\midrule
\rowcolor[gray]{.95} \multicolumn{4}{l}{\textit{Sensitivity Analyses}} \\
25\% & 0.595 $ \pm $ 0.005 & 0.032 $ \pm $ 0.005 & 1.3 $\pm$ 0.1 \\
75\% & 0.612 $ \pm $ 0.006 & 0.015 $ \pm $ 0.006 & 1.2 $\pm$ 0.1 \\
$w_t \ 1.0$ & 0.619 $ \pm $ 0.004 & 0.008 $ \pm $ 0.004 & 1.1 $\pm$ 0.1 \\
$w_t \ 0.7$ & 0.612 $ \pm $ 0.008 & 0.015 $ \pm $ 0.008 & 1.1 $\pm$ 0.1 \\
$w_t \ 1.0 \to 0.3$ & 0.619 $ \pm $ 0.004 & 0.008 $ \pm $ 0.004 & 1.1 $\pm$ 0.1 \\
\midrule
Ours & 0.624 $\pm$ 0.002 & 0.003 $\pm$ 0.002 & 1.1 $\pm$ 0.1 \\
\bottomrule
\end{tabular}
}
\end{table}

The relative ablation pattern persists under the stronger source-disjoint setting; in particular, removing the offline component still produces the largest degradation, which is notable because the historical corpus contains no dataset sharing CasiMedicos-Arg's underlying question source. On the other hand, the penalties associated with the individual ablations are smaller than in the primary LODO setting. Overall, the source-disjoint experiments support the transferability of the two-phase offline-online design, and at the same time illustrate that the relative importance of the online components can vary across target datasets. 

As for sensitivity, the sensitivity to early-stop truncation length and online evaluation budget mirror that of the main LODO setting. In particular, increasing the observed SFT trajectory from 25\% to 50\% substantially improves selection performance, while extending it to 75\% provides no additional benefit. However, there is reduced sensitivity to the exact tempering schedule of the offline guidance. This suggests that the benefit of combining offline guidance with online correction is robust to the weighting schedule on this specific target.

\subsection{Feature Importance Analysis}
\label{appendix:featsanalysis}

We examined normalized XGBoost gain to characterize which feature groups are used by the offline ranker and early-stop regressor. For the ranker, dataset-derived features dominate the analysis, consistent with its purpose of learning how dataset context affects configuration preferences. Potentially, although dataset features are constant within each ranking group, splits on these features can partition datasets into different regions before subsequent splits on configuration features, allowing for different configuration preferences to be learned for different regions of dataset-feature space. For the regressor, the contribution of dataset-derived features falls while training dynamics features make a substantial contribution, consistent with the intended decomposition of AutoPipe: dataset characteristics shape the transferable offline guidance, while observed optimization dynamics provide evidence for target-specific correction. 

\begin{table}[h]
\centering
\small
\caption{Feature importance analyses for the ranker and early stop regressor.} 
\label{tab:feat-importance}
\resizebox{\columnwidth}{!}{%
\begin{tabular}{lcc}
\toprule
\textbf{Feature Category} & \textbf{Gain} & \textbf{Dataset Range} \\
\midrule
\rowcolor[gray]{.95} \multicolumn{3}{l}{\textit{Offline Ranker}} \\
Configuration & 8.87 $\pm$ 0.05 & 8.26 - 9.63 \\
Simple dataset & 10.26 $\pm$ 0.02 & 9.89 - 10.81 \\
Rich dataset & 27.09 $\pm$ 0.20 & 26.25 - 28.08 \\
Dataset overlap & 53.78 $\pm$ 0.24 & 51.72 - 54.74 \\
\midrule
\rowcolor[gray]{.95} \multicolumn{3}{l}{\textit{Early Stop Regressor}} \\
Configuration & 13.10 $\pm$ 0.34 & 11.29 - 14.78 \\
Simple dataset & 6.90 $\pm$ 0.12 & 6.14 - 7.74 \\
Rich dataset & 16.12 $\pm$ 0.30 & 14.54 - 18.36 \\
Dataset overlap & 33.58 $\pm$ 0.45 & 31.49 - 36.61 \\
Training dynamics & 30.30 $\pm$ 0.16 & 28.58 - 31.69 \\
\bottomrule
\end{tabular}
}
\end{table}

\end{document}